\documentclass[11pt]{article}

\usepackage[preprint]{acl}

\usepackage{fontspec}
\usepackage[bidi=default]{babel}
\babelprovide[main, import]{english}
\babelprovide[import]{arabic}
\babelfont[arabic]{rm}[
  Extension = .ttf,
  UprightFont = Amiri-Regular,
  BoldFont = Amiri-Bold,
  ItalicFont = Amiri-Italic,
  BoldItalicFont = Amiri-BoldItalic,
  Script=Arabic
]{Amiri}  
\usepackage{graphicx}
\usepackage{inconsolata}
\usepackage[utf8]{inputenc}
\usepackage[T1]{fontenc}
\usepackage{colortbl}
\usepackage{todonotes}

\usepackage{times}
\usepackage{latexsym}
\usepackage{fontspec}
\usepackage[bidi=default]{babel}

\babelfont[arabic]{rm}[
  Extension = .ttf,
  UprightFont = Amiri-Regular,
  BoldFont = Amiri-Bold,
  ItalicFont = Amiri-Italic,
  BoldItalicFont = Amiri-BoldItalic,
  Script=Arabic
]{Amiri}
\usepackage{microtype}
\usepackage{inconsolata}
\usepackage{xcolor}
\usepackage{booktabs}
\usepackage{multirow}
\usepackage{array}
\usepackage{amsmath}
\usepackage{amssymb}
\usepackage{makecell}
\usepackage{pifont}
\usepackage{float}
\usepackage{subcaption}
\usepackage{tabularx}
\usepackage{latexsym}
\usepackage{xspace}
\usepackage{tipa}
\usepackage[most]{tcolorbox}
\usepackage{listings}
\usepackage{comment}
\usepackage{textcomp}
\usepackage{lipsum}
\usepackage{hyperref}
\usepackage[nameinlink]{cleveref}
\usepackage{fontawesome5}
\usepackage{enumitem}
\setlist[itemize]{leftmargin=14pt}
\usepackage{xcolor}
\definecolor{softgray}{HTML}{EBE0D2}
\usepackage{xcolor}
\definecolor{darkgreen}{rgb}{0.0, 0.5, 0.0}
\usepackage{soul}
\sethlcolor{yellow!35}
\makeatletter
\def\SOUL@hlpreamble{%
  \setul{}{3ex}\let\SOUL@stcolor\SOUL@hlcolor\SOUL@stpreamble}
\makeatother

\newcolumntype{L}[1]{>{\RaggedRight\arraybackslash}p{#1}}
\newcolumntype{Y}{>{\RaggedRight\arraybackslash}X}

\newcommand{\name}{\textsc{Sahm}\xspace}

\title{
    \raisebox{-0.2cm}{\includegraphics[width=1.0cm]{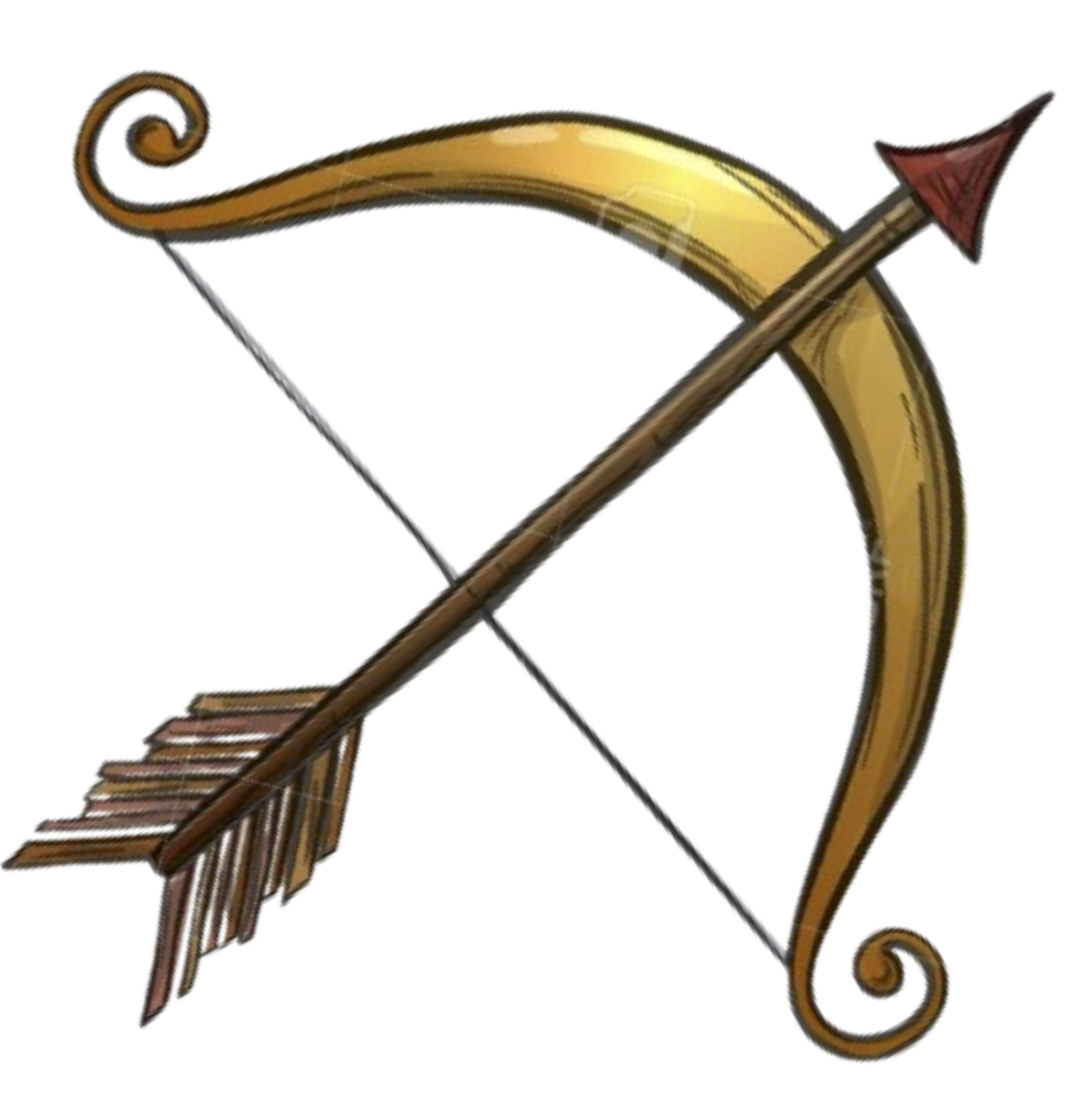}} 
    \hspace{0.01cm} \name : A Benchmark for Arabic Financial \\and Shari’ah-Compliant Reasoning
}

\author{
\textbf{Rania Elbadry}\textsuperscript{1} \
\textbf{Sarfraz Ahmad}\textsuperscript{1} \
\textbf{Ahmed Heakl}\textsuperscript{1} \
\textbf{Dani Bouch}\textsuperscript{1} \
\textbf{Momina Ahsan}\textsuperscript{1} \\
\textbf{Muhra AlMahri}\textsuperscript{1} \
\textbf{Marwa Elsaid Khalil}\textsuperscript{1} \
\textbf{Yuxia Wang}\textsuperscript{2} \\
\textbf{Salem Lahlou}\textsuperscript{1} \
\textbf{Sophia Ananiadou}\textsuperscript{3} \
\textbf{Veselin Stoyanov}\textsuperscript{1} \
\textbf{Jimin Huang}\textsuperscript{4} \\
\textbf{Xueqing Peng}\textsuperscript{4}\thanks{Corresponding author} \
\textbf{Preslav Nakov}\textsuperscript{1} \
\textbf{Zhuohan Xie}\textsuperscript{1} \\
\textsuperscript{1}MBZUAI \
\textsuperscript{2}INSAIT, Sofia University "St. Kliment Ohridski"\\
\textsuperscript{3}The University of Manchester \
\textsuperscript{4}The Fin AI \\
\parbox{\linewidth}{\centering
\texttt{\{rania.elbadry, zhuohan.xie\}@mbzuai.ac.ae} \quad \texttt{xueqing.peng2024@gmail.com}\\
\faGlobe\ \href{https://mbzuai-nlp.github.io/SAHM/}{Project}
\quad
\faDatabase\ \href{https://huggingface.co/SahmBenchmark}{SAHM}
\quad
\faGithub\ \href{https://github.com/mbzuai-nlp/SAHM}{Code}
\quad
\faTrophy\ \href{https://mbzuai-nlp.github.io/SAHM/leaderboard.html}{Leaderboard}
}
}
\begin{document}
\maketitle
\begin{abstract}

English financial NLP has advanced rapidly through benchmarks targeting earnings analysis, market sentiment, tabular reasoning, and financial question answering, yet Arabic financial NLP remains virtually nonexistent, despite 422 million speakers, \$4.9 trillion in Gulf sovereign wealth, and a \$4--5 trillion Islamic finance industry requiring specialized Shari'ah compliance over instruments like sukuk, murabaha, and takaful. We introduce \textsc{Sahm}, the first Arabic financial benchmark spanning seven tasks: AAOIFI standards QA, fatwa-based QA/MCQ, accounting and business exams, financial sentiment analysis, extractive summarization, and event-cause reasoning, comprising 14,380 expert-verified instances from authentic regulatory, juristic, and corporate sources. Evaluating 20 LLMs, we find Arabic fluency does not imply financial reasoning: models achieving 91\% on recognition tasks drop sharply on generation, and event-cause reasoning exposes the widest performance gap (1.89--9.84/10). We release the benchmark and dataset to support trustworthy Arabic financial assistants.

\end{abstract}
\section{Introduction}

\begin{figure*}[t]
    \centering
    \includegraphics[width=1.0\linewidth]{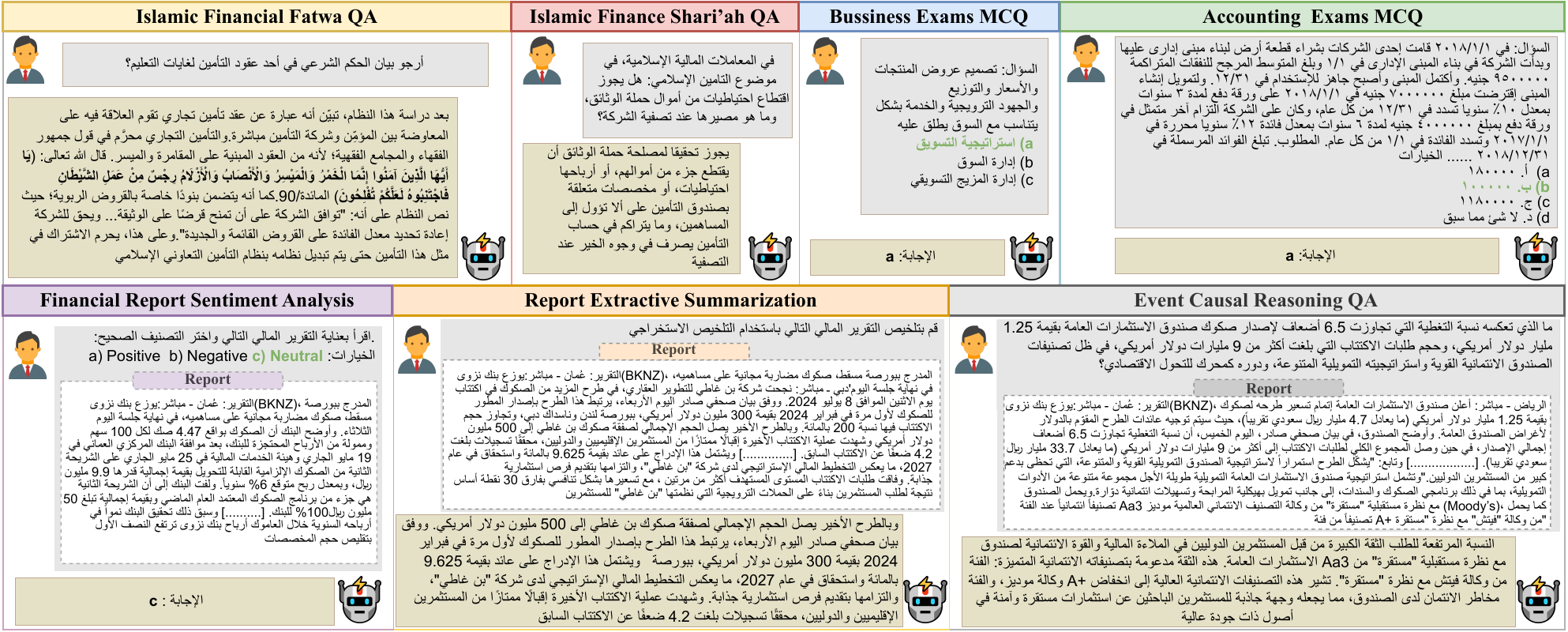} 
    \caption{\textbf{Examples of the diverse tasks included in \name{}}, covering juristic Q\&A, business and accounting MCQs, financial sentiment analysis, report summarization, \& event causal reasoning.}
    \label{fig:overview}    
    \vspace{-1.0em}
\end{figure*}

The Gulf Cooperation Council (GCC) generates large volumes of Arabic financial text, including central bank reports, regulatory filings, corporate disclosures, and \textit{fatwas} that provide jurisprudential rulings. Despite this, evaluation of Large Language Models (LLMs) on Arabic financial content remains limited.  
English financial NLP has advanced rapidly through dedicated benchmarks~\citep{fiqa,tatqa,finqa,convfinqa,financemath,finchain}, with multilingual extensions for other languages~\citep{cfinbench,fit-es,plutus,multifinben}.

Arabic benchmarks remain limited in scope: ArBanking77~\citep{arbanking77} addresses only banking intent, and Arabic-centric LLMs~\citep{jais,falcon-arabic,ain,fanar} have not been evaluated on financial domains. Islamic finance further illustrates this gap. Unlike conventional finance, it requires Shari'ah review guided by standards issued by AAOIFI.\footnote{\url{https://aaoifi.com}} Although resources such as Fatwaset~\citep{fatwaset} and Hajj FQA~\citep{hajjfqa} exist, they focus on general juristic QA rather than financial reasoning. As a result, LLMs remain untested on tasks that combine legal and financial analysis.

We introduce \name{}, the first Arabic financial NLP benchmark unifying modern finance and Islamic jurisprudence, two high-stakes domains shaping trillions in assets yet missing from LLM evaluation. This enables joint evaluation of compliance and financial reasoning. \name{} spans seven expert-verified tasks grounded in AAOIFI standards, fatwa archives from seven countries, and corporate disclosures (Figure~\ref{fig:overview}). Evaluating 20 LLMs reveals that Arabic fluency does not guarantee financial reasoning: base Arabic models rank in the bottom 25\% despite being designed for Arabic. However, fine-tuning on \name{} closes this gap: domain-adapted models gain up to +26 points on Accounting and +25 points on Business, enabling 7--8B models to surpass GPT-5 and match 72B open-source baselines. Our contributions:

\begin{itemize}
\item The first Arabic finance benchmark (14,380 instances; 7 tasks) jointly evaluating Shari'ah-compliant reasoning (fatwa QA, Islamic finance standards) and core financial competencies (accounting MCQ, sentiment, event-cause QA), addressing a major resource gap for Arabic financial NLP.
\item A comprehensive benchmark of 20 LLMs showing that Arabic fluency does not guarantee financial reasoning: models that score up to 91\% on MCQ-style tasks degrade substantially on open-ended generation, with the largest gap on Event--Cause QA (1.89--9.84/10).
\item Evidence that targeted adaptation rivals scale for Arabic financial NLP: fine-tuning on \name{} yields two complementary 7--8B models \textsc{Sahm-ALLAM-7B} (peak accuracy, surpassing GPT-5 by +21.3 points on Business MCQ, 93.99\% vs.\ 72.68\%) and \textsc{Sahm-Jais-8B} (uniformly positive transfer across all tasks) while matching 72B open-source baselines on average demonstrating $\sim$10$\times$ parameter efficiency and establishing domain adaptation as a practical, cost-effective route to trustworthy Arabic financial assistants where frontier API access may be limited.
\end{itemize}

\begin{figure*}[t]
    \centering
    \includegraphics[width=1\linewidth]{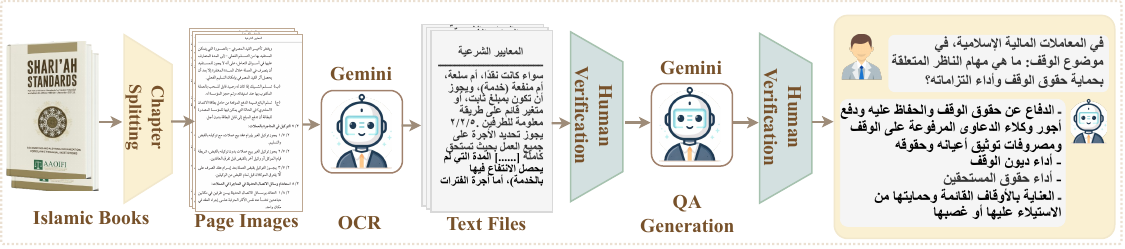} 
    \caption{\textbf{Pipeline for constructing the Islamic Finance Shari'ah Standards QA dataset.} A hybrid LLMs-human pipeline converts AAOIFI standards into QA pairs through OCR and generation stages, each followed by expert verification to ensure linguistic accuracy and legal fidelity.}
    \label{fig:islamicstandard}
\end{figure*}

\section{Related Work}

\paragraph{Financial NLP Benchmarks:} English financial NLP has matured through progressively challenging benchmarks. Early work focused on classification and extraction~\citep{araci2019finbert}, while recent datasets target numerical reasoning over tables (FinQA~\citep{finqa}, TAT-QA~\citep{tatqa}), multi-turn dialogue (ConvFinQA~\citep{convfinqa}), and chain-of-thought verification (FinChain~\citep{finchain}). Comprehensive suites such as FinBen~\citep{finben} and PIXIU~\citep{pixiu} now span 24 tasks including sentiment, NER, and argument mining. 

Multilingual extensions have emerged for Chinese (CFinBench~\citep{cfinbench}), and Greek (Plutus~\citep{plutus}), demonstrating that culturally grounded evaluation reveals failure modes invisible in English-only testing. Yet Arabic, spoken by 422M people across economies managing \$4.9T in sovereign wealth~\citep{alhajraf2025swf}, lacks any comparable financial benchmark.

\paragraph{Arabic NLP and the Evaluation Gap:} Arabic resources have grown substantially, but remain shallow in financial coverage. ArBanking77~\citep{arbanking77} addresses banking intent detection; Fatwaset~\citep{fatwaset} and Hajj-FQA~\citep{hajjfqa} target religious QA. These datasets support general understanding but not reasoning for compliance, numerical analysis, or Shari'ah-aligned decisions. This gap is critical in high-stakes financial settings, where incorrect reasoning can lead to regulatory violations or financial loss. Moreover, the absence of targeted benchmarks limits our ability to diagnose and improve model performance in real-world Arabic financial applications. This gap is significant as Arabic financial texts present distinct challenges: mixed numeral systems (Eastern \foreignlanguage{arabic}{٠١٢٣} and Western 0123), code-switching with English acronyms (IFRS, AAOIFI), and domain-specific terminology from Islamic jurisprudence (\textit{riba}, \textit{gharar}, \textit{sukuk}). Meanwhile, Arabic-centric LLMs, including Jais~\citep{jais}, Falcon-Arabic~\citep{falcon-arabic}, AIN~\citep{ain}, and Fanar~\citep{fanar}, are evaluated only on generic benchmarks that ignore these complexities.

\section{\name{}}
\label{sec:sahm}
\begin{table*}[t]
\centering
\LARGE
\resizebox{\textwidth}{!}{%
\begin{tabular}{l l c c c c c}
\toprule
\rowcolor{softgray}
\textbf{Task} & \textbf{Dataset} & \textbf{N} & \textbf{Avg.\ Words (Input)} & \textbf{Avg.\ Chars (Input)} & \textbf{Avg.\ Words (Answer)} & \textbf{Avg.\ Chars (Answer)} \\
\midrule
\multirow{4}{*}{MCQ}
 & Accounting Exams MCQ                    & 167   & 111.5 $\pm$ 91.1   & 674.3 $\pm$ 550.5    & 1.0 $\pm$ 0.0   & 1.0 $\pm$ 0.0 \\
 & Business Exams MCQ                      & 183   & 46.3  $\pm$ 12.2   & 298.3 $\pm$ 71.6     & 1.0 $\pm$ 0.0   & 1.0 $\pm$ 0.0 \\
 & Islamic Financial Fatwa MCQ             & 2{,}000 & 93.1 $\pm$ 14.7  & 536.7 $\pm$ 82.6     & 1.0 $\pm$ 0.0   & 1.0 $\pm$ 0.0 \\
 & Financial Report Sentiment Analysis MCQ & 80    & 292.3 $\pm$ 139.3  & 1{,}780.7 $\pm$ 841.9 & 1.0 $\pm$ 0.0  & 1.0 $\pm$ 0.0 \\
\midrule
\multirow{4}{*}{Open-Ended}
 & Event--Cause Reasoning QA               & 80    & 413.6 $\pm$ 299.9  & 2{,}503.7 $\pm$ 1{,}752.1 & 350.6 $\pm$ 101.8 & 2{,}170.4 $\pm$ 635.8 \\
 & Islamic Fatwa QA                        & 2{,}000 & 64.1 $\pm$ 36.2  & 377.3 $\pm$ 200.6    & 89.9  $\pm$ 58.4  & 492.5 $\pm$ 324.0 \\
 & Islamic Shar\={\i}\textquotesingle a Standards QA & 811 & 140.1 $\pm$ 5.2 & 287.0 $\pm$ 39.5 & 33.2 $\pm$ 22.0 & 192.1 $\pm$ 129.8 \\
 & Report Extractive Summarization         & 80    & 355.4 $\pm$ 165.4  & 2{,}144.3 $\pm$ 972.3 & 157.4 $\pm$ 66.5 & 929.1 $\pm$ 391.7 \\
\bottomrule
\end{tabular}
}
\caption{\textbf{Dataset statistics for \name{}.} Mean $\pm$ standard deviation of word and character counts per instance, computed over the test split of each dataset. For MCQ tasks the answer is a single letter (A--D), hence the constant 1.0 word/char count.}
\label{tab:dataset_stats}
\vspace{-0.6em}
\end{table*}

We introduce \name{}, a comprehensive benchmark for evaluating Arabic financial reasoning across diverse, real-world tasks spanning Islamic finance, accounting, and market analysis. The benchmark is designed to capture both rule-based reasoning grounded in Shari’ah standards and applied financial understanding in authentic Arabic contexts. It covers multiple task formats, including question answering, multiple-choice reasoning, sentiment analysis, and summarization, enabling holistic assessment of model capabilities. Table~\ref{tab:dataset_stats} provides an overview of dataset composition, task distribution, and train--test splits.

\subsection{Islamic Finance Shari’ah Standards QA}

Finance in the Gulf and the wider MENA region differs from Western systems: banks, insurers, and capital markets must comply with Islamic principles governed by detailed Shari’ah standards. Frameworks such as \textit{AAOIFI} and local regulations specify how financial instruments are structured, e.g., lease-to-own arrangements in \emph{Ijara} (\foreignlanguage{arabic}{إجارة}) and compliance requirements for Sukuk\footnote{\foreignlanguage{arabic}{صكوك} (sukuk) are Shari’ah-compliant financial certificates representing ownership in underlying assets rather than interest-bearing debt.} issuance~\citep{aaoifi,ifsb2024,sama}. Yet, most financial benchmarks assume Western instruments (e.g., interest-bearing loans, conventional bonds), leaving models untested on region-specific reasoning about contract permissibility, legal constraints, and Shari’ah compliance. To address this gap, we construct the first Islamic Shari’ah Standards QA dataset from the $1{,}264$-page AAOIFI compendium spanning $52$ standards, enabling systematic evaluation of rule-based Islamic financial reasoning.

We built the dataset through a multi-step pipeline that converts the AAOIFI compendium into text via OCR with \texttt{Gemini-2.5-Pro}~\citep{gemini2.5pro_google} (Appendix~\ref{sec:ocr-eval} provides details) recommended by \citet{heakl2025kitab}. Two Islamic finance experts verified the text, preserving diacritics, numerals, and domain-specific terms. In a review of a $25\%$ sample, experts measured a high exact-match rate of $98.7\pm0.7\%$ (95\% CI) and strong inter-annotator agreement ($\kappa = 0.962$), confirming OCR reliability. The remaining $1.3\%$ mismatches were minor orthographic or formatting issues (e.g., spacing, punctuation, diacritics), which were corrected in the canonical text; no errors altered the substance of any Shari’ah ruling. After cleanup, we grouped the verified text into thematic clusters (e.g., Muraba\d{h}a) and used \texttt{Gemini-2.5-Pro} to draft candidate Arabic question–answer pairs. Domain experts refined and validated the samples to ensure each question captured the correct ruling with all conditions and exceptions. This human-in-the-loop pipeline transforms dense regulatory prose into high-quality, legally faithful QA pairs for benchmarking Shari’ah-compliant financial reasoning (Figures~\ref{fig:overview},~\ref{fig:islamicstandard}).

\subsection{Islamic Financial Fatwa QA}

We scraped fatwā archives from $13$ official websites across $7$ Arab countries to capture the breadth of real-world financial questions Muslims ask (Table~\ref{tab:fatwa_websites}). The initial crawl yielded $20k$ fatwas, which we cross-checked against the public FatwaSet~\citep{fatwaset} to remove duplicates and then organized into 11 finance-related categories (Table~\ref{tab:category_counts}), including \foreignlanguage{arabic}{زكاة} (almsgiving), \foreignlanguage{arabic}{ربا} (usury), and \foreignlanguage{arabic}{مرابحة} (cost-plus financing). We then transformed these long, formal texts into concise QA pairs via \texttt{Gemini-2.5-pro} while preserving their juristic meaning.

Specifically, we removed introductory invocations (e.g., \foreignlanguage{arabic}{``الحمد لله، والصلاة والسلام على رسول الله''}) and rhetorical openers (e.g., \foreignlanguage{arabic}{``أما بعد''}) to expose the core inquiry and ruling. We stripped HTML artifacts and redundant navigational references while retaining key metadata such as source URLs for traceability. This normalization step reduces noise and standardizes inputs for downstream QA construction. It also ensures that models focus on substantive legal content rather than stylistic variations. This pipeline removes greetings, honorifics, hyperlinks, and scholar names while preserving Qurʾānic citations, juristic terminology, and legal reasoning. Further details in Appendix~\ref{sec:fatwa_prompt}. Two native Arabic speakers manually reviewed 10\% of the normalized data from each category to verify clarity, linguistic fidelity, and domain correctness. This process resulted in exactly $9{,}953$ high-quality training samples and $2{,}000$ held-out finance-focused test cases (Figure~\ref{fig:overview}).

Afterwards, we converted each test QA pair into multiple-choice (MCQ) format via \texttt{Gemini-2.5-Pro}, enabling both open-ended fatwā reasoning and recognition-style testing. Each MCQ consists of one correct answer derived from the source fatwā and three plausible distractors reflecting common misconceptions. Two native Arabic annotators independently reviewed the test set to assess MCQ correctness, alignment with the source fatwā, and distractor plausibility. The annotators achieved high agreement (Cohen's $\kappa = 0.89$). Following this pilot phase, we conducted a calibration round in which annotators discussed disagreements, resolved ambiguous cases, and refined shared labeling criteria. One annotator validated the remaining MCQs, ensuring alignment with source fatwās, correct terminology, and no misleading options. A final audit confirmed that $95\%$ of MCQs aligned exactly with their original QA pairs; we discarded the remaining $5\%$ and excluded them from evaluation.

\subsection{Business \& Accounting Exams MCQ}
Professional accounting assessment resources remain largely English-centric, with key certifications such as the CPA exam conducted exclusively in English. To address this gap and the limited availability of Arabic training materials despite the existence of IFRS translations, we design culturally and linguistically adapted MCQ samples covering IFRS treatments, financial ratios, budgeting, and costing, incorporating authentic Arabic financial terminology such as \foreignlanguage{arabic}{معدل دوران الأصول} (asset turnover ratio) and \foreignlanguage{arabic}{زكاة الشركات} (corporate almsgiving) within contextually accurate scenarios rather than direct translations of Western exam questions (Appendix~\ref{app:exam_prompts}).
We constructed the dataset by collecting 10 business exams and 8 accounting exams from multiple Arabic-speaking countries. We extract the text from the exam PDFs via \texttt{Gemini-2.5-Pro} following \citet{heakl2025kitab}, after which two native Arabic-speaking annotators reviewed by comparing the OCR output against the original questions, correcting recognition errors, and validating formatting. The final dataset contains $457$ business questions and $416$ accounting questions, examples in Figure~\ref{fig:overview}.

\subsection{Financial Report Sentiment Analysis}

Despite managing trillions in assets, Arabic markets lack region-specific sentiment benchmarks. Existing English datasets~\citep{maia2018fiqa} focus on Western market narratives and do not capture signals central to MENA markets, including OPEC+ production decisions, \foreignlanguage{arabic}{صكوك} (sukuk) issuances, subsidy reforms, and Shari’ah-compliance rulings. These challenges are amplified by culturally grounded terminology, e.g., \foreignlanguage{arabic}{مرابحة} (cost-plus financing), and stylistic variation in Arabic reporting, where subtle modifiers can reverse sentiment polarity. To address this gap, We construct the first Arabic financial sentiment benchmark from authentic market reports.

We collect $200$ Arabic financial reports, 100 Islamic finance–focused and 100 general, from Argaam\footnote{\url{https://www.argaam.com/}}, and annotate them with three document-level sentiment labels: \texttt{Positive}, \texttt{Negative}, and \texttt{Neutral}. Two native Arabic annotators labeled all reports using a custom web-based platform (Figure~\ref{fig:sentiment_annotation}) following guidelines emphasizing holistic document interpretation, achieving high agreement ($\kappa = 0.91$). We then conducted a calibration phase to resolve disagreements and refine criteria. For mixed-signal reports, we assign the dominant polarity if >$60\%$ of content supports it; otherwise \texttt{Neutral}, with a third expert adjudicating residual disagreements. The dataset is split into $120$ training and $80$ test reports.

\subsection{Report Extractive Summarization}
Extractive summarization is critical for Arabic financial reporting, where annual reports are written in Arabic but frequently contain mixed numeral systems, embedded English financial acronyms and brand names rendered in Arabic script (e.g., \foreignlanguage{arabic}{المعايير الدولية للتقارير المالية} / IFRS and \foreignlanguage{arabic}{إتش إس بي سي} / HSBC), and specialized Islamic finance terminology such as \foreignlanguage{arabic}{صكوك} (sukuk). Misinterpreting or omitting these elements can distort regulatory interpretation, compliance assessment, and financial valuation. To support this task, we compile $200$ Arabic financial reports, $100$ general and $100$ Islamic from Argaam and annotate them with extractive summaries written in Arabic by two native Arabic speakers. Rather than treating summarization as a subjective agreement task, we use ROUGE~\citep{rouge} to measure overlap between independently produced summaries as a consistency check and select the more complete summary as the gold reference. We split the dataset into $120$ training reports and $80$ test reports. Further details in Appendix~\ref{sec:report-extractive-summarization}.

\subsection{Event-Cause Reasoning QA}

Financial event-cause reasoning is underexplored in Arabic due to the lack of datasets that require models to explain why financial events occur and what implications they entail. To address this gap, we introduce an event-cause reasoning task that evaluates whether models can analyze Arabic financial reports and produce analytical explanations grounded in reported financial data, including market movements and \foreignlanguage{arabic}{صكوك} issuances. 

We collect $200$ Arabic financial reports ($100$ Islamic, $100$ general) from \textit{Argaam}. Two native Arabic financial experts annotate each report by creating one analytical question linking multiple data points and a concise answer explaining causes and implications using only the article content; a pilot on $20$ reports ensures guideline clarity. Details are in Appendix~\ref{app:event_cause_guidelines}. We assess quality via Cohen’s $\kappa = 0.86$ for event-cause identification and ROUGE overlap for answer consistency. After calibration to resolve disagreements, one expert completes the remaining annotations under the agreed criteria.

\begin{table*}[t]
\centering
\resizebox{\textwidth}{!}{%
\begin{tabular}{l *{5}{r} c *{4}{r}}
\toprule
\multirow{3}{*}{\textbf{Model}} &
\multicolumn{5}{c}{\textbf{MCQ (Accuracy \% $\uparrow$)}} &&
\multicolumn{4}{c}{\textbf{Open-Ended QA (Score 0--10 $\uparrow$)}} \\
\cmidrule{2-6} \cmidrule{8-11}
& \multicolumn{4}{c}{\textbf{Datasets}} & \textbf{} &&
\multicolumn{3}{c}{\textbf{Datasets}} & \textbf{} \\
\cmidrule(lr){2-5}\cmidrule(lr){6-6}\cmidrule(lr){8-10}\cmidrule(lr){11-11}
& \textbf{Accounting} & \textbf{Business} & \textbf{Fatwā} & \textbf{Sentiment} & \textbf{Mean} &&
\textbf{Event-Cause QA} & \textbf{Islamic-Standards-QA} & \textbf{Fatwa-QA} & \textbf{Mean} \\
\midrule
\rowcolor{softgray}
\multicolumn{11}{c}{\textbf{Open-source Models: $\geq$ 70B Parameters}} \\
\midrule
Qwen2.5-72B-Instruct~\cite{qwen25}
& 65.87$_{\pm 2.70}$ & 74.86$_{\pm 0.32}$ & 84.65$_{\pm 0.33}$ & 75.00$_{\pm 1.25}$ & 75.10 &&
8.1000$_{\pm 0.10}$ & 5.6330$_{\pm 0.10}$ & 5.3912$_{\pm 0.06}$ & 6.3747 \\
LLaMA-3.1-70B~\cite{llama3}
& 52.10$_{\pm 2.79}$ & 77.60$_{\pm 1.14}$ & 84.90$_{\pm 0.15}$ & 80.00$_{\pm 3.31}$ & 73.65 &&
6.623$_{\pm 0.15}$ & 3.7245$_{\pm 0.10}$ & 4.7607$_{\pm 0.08}$ & 5.036 \\
\midrule
\rowcolor{softgray}
\multicolumn{11}{c}{\textbf{Open-source Models: $<$ 70B Parameters}} \\
\midrule
Qwen2.5-14B-Instruct~\cite{qwen25}
& 49.10$_{\pm 3.93}$ & 63.39$_{\pm 0.83}$ & 76.05$_{\pm 0.85}$ & 57.50$_{\pm 3.82}$ & 61.51 &&
7.4975$_{\pm 0.10}$ & 4.8806$_{\pm 0.10}$ & 4.0576$_{\pm 0.06}$ & 5.4786 \\
Qwen2.5-7B-Instruct~\cite{qwen25}
& 48.50$_{\pm 2.85}$ & 59.56$_{\pm 1.14}$ & 70.00$_{\pm 0.28}$ & 55.00$_{\pm 1.91}$ & 58.27 &&
6.1038$_{\pm 0.12}$ & 3.4039$_{\pm 0.10}$ & 2.6815$_{\pm 0.08}$ & 4.0631 \\
Gemma-2-9B-IT~\cite{gemma2}
& 49.10$_{\pm 2.74}$ & 63.39$_{\pm 3.83}$ & 66.60$_{\pm 0.61}$ & 55.00$_{\pm 1.44}$ & 58.52 &&
7.1438$_{\pm 0.08}$ & 4.2306$_{\pm 0.08}$ & 3.4266$_{\pm 0.06}$ & 4.9336 \\
Gemma-3-27B-IT~\cite{gemma3}
& 53.89$_{\pm 2.16}$ & 73.22$_{\pm 0.32}$ & 80.65$_{\pm 0.18}$ & 80.00$_{\pm 0.72}$ & 71.94 &&
8.7188$_{\pm 0.05}$ & 6.1708$_{\pm 0.08}$ & 5.1929$_{\pm 0.05}$ & 6.6942 \\
Gemma-3-4B-IT~\cite{gemma3}
& 38.32$_{\pm 2.27}$ & 67.76$_{\pm 0.32}$ & 61.35$_{\pm 0.18}$ & 75.00$_{\pm 1.44}$ & 60.61 &&
7.4075$_{\pm 0.08}$ & 2.8985$_{\pm 0.08}$ & 2.4767$_{\pm 0.06}$ & 4.2609 \\
LLaMA-3.1-8B~\cite{llama3}
& 41.92$_{\pm 3.28}$ & 60.66$_{\pm 4.45}$ & 64.05$_{\pm 3.62}$ & 73.75$_{\pm 5.77}$ & 60.60 &&
4.9231$_{\pm 0.18}$ & 2.5168$_{\pm 0.12}$ & 1.4025$_{\pm 0.08}$ & 2.9475 \\
Mixtral-8x7B-Instruct~\cite{jiang2024mixtralexperts}
& 32.93$_{\pm 1.04}$ & 60.66$_{\pm 0.63}$ & 62.15$_{\pm 0.34}$ & 70.00$_{\pm 0.72}$ & 56.44 &&
4.5538$_{\pm 0.08}$ & 2.4980$_{\pm 0.08}$ & 1.7896$_{\pm 0.06}$ & 2.9471 \\
\midrule
\rowcolor{softgray}
\multicolumn{11}{c}{\textbf{Proprietary Models: Reasoning-Enhanced}} \\
\midrule
GPT-5~\cite{gpt5}
& 65.27$_{\pm 2.27}$ & 72.68$_{\pm 1.26}$ & 90.75$_{\pm 0.45}$ & 78.75$_{\pm 1.25}$ & 76.86 &&
9.6831$_{\pm 0.03}$ & 8.7965$_{\pm 0.05}$ & 8.0515$_{\pm 0.04}$ & 8.8437 \\
GPT-4o~\cite{gpt4o}
& 60.48$_{\pm 2.07}$ & \textbf{78.14}$_{\pm 0.32}$ & 87.70$_{\pm 0.10}$ & 77.50$_{\pm 0.00}$ & 75.96 &&
8.3125$_{\pm 0.06}$ & 6.6598$_{\pm 0.08}$ & 6.5219$_{\pm 0.04}$ & 7.1647 \\
\midrule
\rowcolor{softgray}
\multicolumn{11}{c}{\textbf{Proprietary Models: General-Purpose}} \\
\midrule
Claude-Opus-4.5~\cite{claude45opus}
& 77.84$_{\pm 2.42}$ & 76.50$_{\pm 1.14}$ & 91.75$_{\pm 0.33}$ & 75.00$_{\pm 2.50}$ & 80.27 &&
9.6818$_{\pm 0.03}$ & 8.0438$_{\pm 0.05}$ & 8.8090$_{\pm 0.03}$ & 8.8449 \\
Claude-Sonnet-4.5~\cite{claude45sonnet}
& 78.44$_{\pm 1.20}$ & 76.50$_{\pm 1.45}$ & 88.15$_{\pm 0.38}$ & 77.50$_{\pm 1.25}$ & 80.15 &&
9.3388$_{\pm 0.04}$ & 8.2588$_{\pm 0.05}$ & 7.6049$_{\pm 0.03}$ & 8.4008 \\
Claude-Haiku-4.5~\cite{claude45haiku}
& 67.66$_{\pm 1.80}$ & 73.77$_{\pm 1.30}$ & 84.90$_{\pm 0.40}$ & 77.50$_{\pm 1.80}$ & 75.96 &&
9.1050$_{\pm 0.05}$ & 7.0002$_{\pm 0.07}$ & 6.5341$_{\pm 0.05}$ & 7.5464 \\
Gemini-3-Flash (preview)~\cite{gemini3flash}
& 76.05$_{\pm 1.95}$ & 74.86$_{\pm 0.95}$ & 89.90$_{\pm 0.30}$ & 81.25$_{\pm 1.25}$ & 80.52 &&
\textbf{9.8369}$_{\pm 0.02}$ & \textbf{9.1649}$_{\pm 0.03}$ & \textbf{9.1571}$_{\pm 0.02}$ & \textbf{9.0798} \\
GPT-4o-mini~\cite{gpt4o}
& 58.08$_{\pm 2.10}$ & 77.60$_{\pm 0.40}$ & 81.75$_{\pm 0.20}$ & 75.00$_{\pm 0.50}$ & 73.61 &&
7.9613$_{\pm 0.08}$ & 5.6094$_{\pm 0.10}$ & 5.3087$_{\pm 0.06}$ & 6.2931 \\
\midrule
\rowcolor{softgray}
\multicolumn{11}{c}{\textbf{Arabic Models}} \\
\midrule
ALLAM-7B~\cite{bari2024allamlargelanguagemodels}
& 44.91$_{\pm 3.55}$ & 68.31$_{\pm 3.83}$ & 74.40$_{\pm 2.83}$ & 58.75$_{\pm 2.00}$ & 61.59 &&
6.8875$_{\pm 0.10}$ & 4.9364$_{\pm 0.08}$ & 4.2185$_{\pm 0.05}$ & 5.3475 \\
Fanar-1-9B~\cite{fanar}
& 47.31$_{\pm 2.42}$ & 66.12$_{\pm 1.67}$ & 74.45$_{\pm 0.35}$ & 58.75$_{\pm 2.60}$ & 61.66 &&
7.5850$_{\pm 0.10}$ & 4.9607$_{\pm 0.08}$ & 4.4600$_{\pm 0.06}$ & 5.6686 \\
SILMA-9B~\cite{silma-9b-2024}
& 50.90$_{\pm 21.73}$ & 69.40$_{\pm 6.61}$ & 62.55$_{\pm 5.57}$ & 30.00$_{\pm 3.75}$ & 53.21 &&
1.8969$_{\pm 0.20}$ & 3.3547$_{\pm 0.12}$ & 2.0711$_{\pm 0.08}$ & 2.4409 \\
Jais-2-8B~\cite{jais}       
& 35.33$_{\pm 3.00}$ & 60.30$_{\pm 2.80}$ & 66.10$_{\pm 1.80}$ & 46.25$_{\pm 2.50}$ & 52.00 &&
4.6922$_{\pm 0.15}$ & 4.245$_{\pm 0.10}$ & 2.5147$_{\pm 0.08}$ & 3.8133 \\
\bottomrule
\end{tabular}%
}
\caption{\textbf{Unified leaderboard comparing MCQ tasks (Accuracy \%) and open-ended QA tasks (Score 0--10).} Values shown as mean$_{\pm\text{std}}$ over 3 runs; open-ended scores are judged by two independent LLM judges. Open-ended QA Mean is averaged over Event-Cause QA, Islamic-Standards-QA, and Fatwa-QA.}
\label{modelperformance}
\end{table*}
\section{Experiments}

\paragraph{Evaluated Models:} We evaluated \textbf{20} models spanning Arabic-centric models~\cite{bari2024allamlargelanguagemodels,fanar,silma-9b-2024} (publicly available instruction-tuned systems for regional adaptation), open-weight models~\cite{gemma2,gemma3,llama3,qwen25,jiang2024mixtralexperts} (strong multilingual and general-purpose baselines), and proprietary models~\cite{gpt4o,gpt5,claude45opus,claude45sonnet,claude45haiku,gemini3flash}, enabling controlled analysis across language, scale, and capability dimensions. To assess whether domain-specific fine-tuning can close the gap between Arabic-centric and frontier models, we fine-tune three Arabic LLMs (ALLAM-7B, Jais-2-8B, and SILMA-9B) on the \name{} training split using LoRA ($r$=64, $\alpha$=128, lr=2e-4, 3 epochs). Detailed model specifications are provided in Table~\ref{tab:models}.

We evaluate Accounting Exams, Business Exams, Fatwa MCQ, and Financial Sentiment with exact-match accuracy, normalizing free-form outputs (e.g., option text/letters) to a single choice before scoring (Appendix~\ref{app:mcq_normalization}). For extractive summarization, we report ROUGE-F1 (ROUGE-1/2/L) against gold extractive references (models are instructed to output verbatim sentences). For Fatwa QA, Shari’ah Standards QA, and Event-Cause QA, we use \texttt{Gemini-2.5-Flash} as an LLM-as-a-judge (blind to model identity). Given the Arabic prompt, gold reference, and model answer, it returns a JSON-validated additive $[0,10]$ score based on a shared rubric assessing alignment with the reference ruling or conclusion, preservation of key constraints or quantitative fidelity, correctness (doctrinal, factual, or financial), Arabic clarity, and grounding. 

We validate the judge with two expert Arabic annotators on 200 randomly sampled outputs across the three tasks (MSE 0.41, Pearson $r$=0.92; inter-annotator agreement $\kappa$=0.84 on discretized scores; Appendix~\ref{app:judge_rubrics}). All judge and model generations use greedy decoding (temperature 0; no sampling) with fixed maximum lengths; full prompts, rubrics, schema, critical checks, and settings appear in Appendix~\ref{app:judge_rubrics}. We organize our findings around three core questions: (1) How do models perform across recognition versus generation tasks? (2) What distinguishes strong Arabic financial reasoning from mere language fluency? (3) Where do models systematically fail, and why?

\begin{figure}[t]
    \centering
    \includegraphics[width=\linewidth]{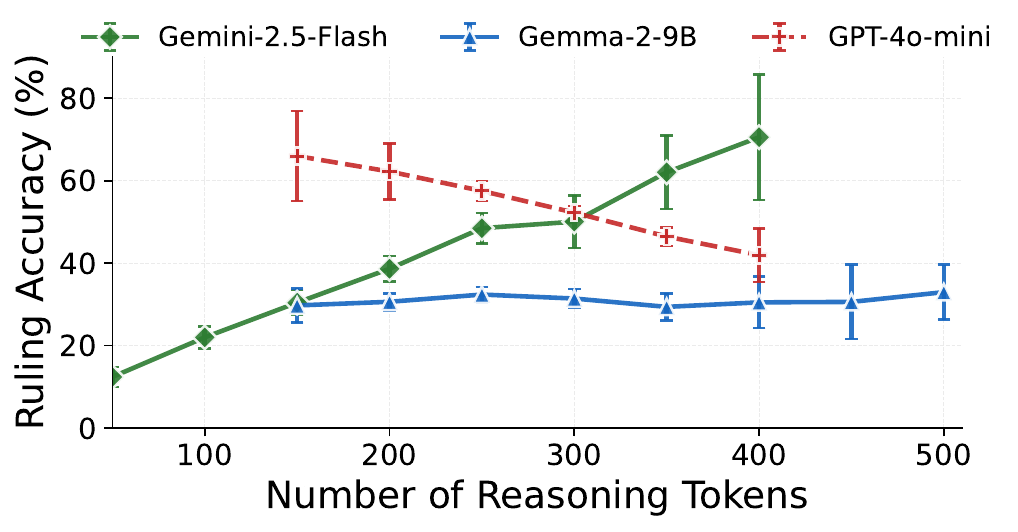}
    \vspace{-2.0em}
    \caption{\textbf{Effect of reasoning token budget on ruling accuracy.} Green indicates improvement with increased budget, red indicates decline, and blue indicates no change.}
    \label{fig:reasoning_tokens}
    \vspace{-1.5em}
\end{figure}

\subsection{Main Results}

Table~\ref{modelperformance} summarizes performance across all tasks, revealing clear disparities across models. We analyze these patterns to identify where models succeed and fail.

\begin{figure*}[t!]
    \centering
    \includegraphics[width=\textwidth]{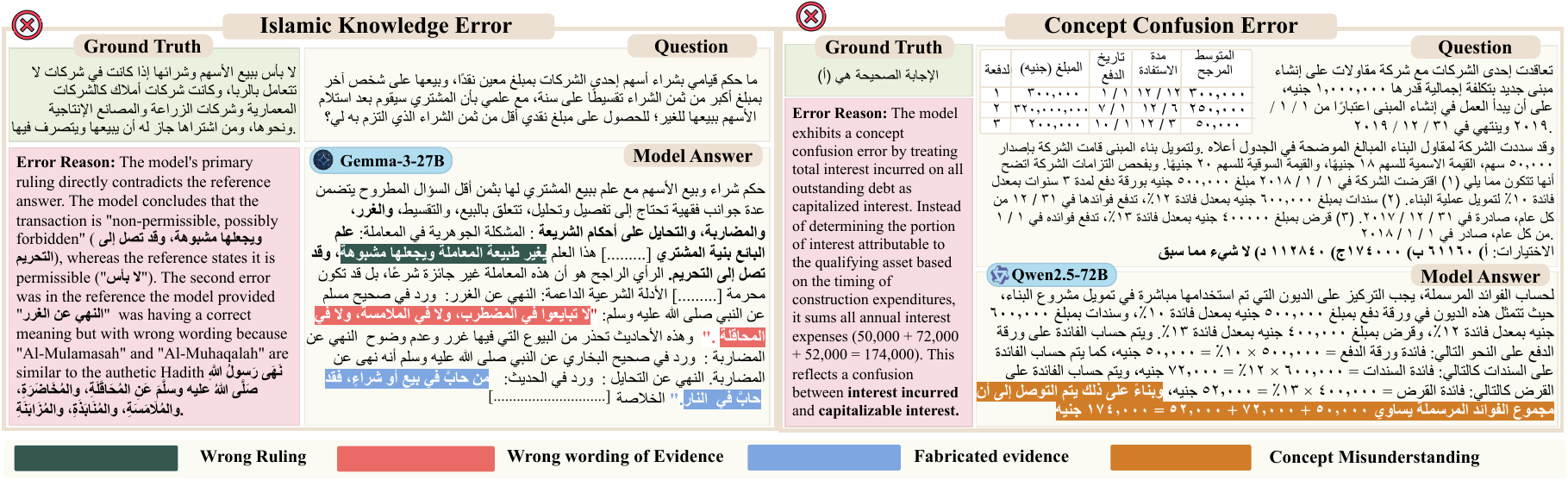}
    \caption{\textbf{Qualitative error analysis showing representative failure modes.} \textit{Left:} Islamic knowledge error where Gemma-3-27B incorrectly rules a permissible transaction as forbidden, citing fabricated evidence with wrong wording of authentic Hadith. \textit{Right:} Concept confusion error where Qwen2.5-72B conflates total interest incurred with capitalizable interest in a construction loan scenario.}
    \label{fig:error_qualitative}
    \vspace{-1.0em}
\end{figure*}

\textbf{Accounting Reasoning Gap:} Shown in Table~\ref{modelperformance}, Claude models exhibit substantial superiority on Accounting tasks, with Claude-Sonnet-4.5 exceeding GPT-5 by over $13\%$ the largest proprietary-to-proprietary gap in our evaluation. Crucially, this disparity cannot be attributed to general Arabic language proficiency alone, as these models achieve near-parity on Business ($76.50\%$ vs. $72.68\%$) and Fatwa ($91.75\%$ vs. $90.75\%$) tasks. We instead attribute this divergence to Claude's stronger capacity for procedural numerical reasoning, the ability to apply rule-based standards (e.g., IFRS, Egyptian Auditing Standards) through multi-step logical chains. This suggests Arabic domain reasoning is distinct from general language proficiency, warranting further study. Notably, Gemini-3-Flash inverts the recognition–generation tradeoff, achieving top Open-Ended QA despite moderate MCQ performance, likely due to longer reasoning chains. This is supported by Figure~\ref{fig:reasoning_tokens}, where the Gemini family shows increased ruling accuracy with larger reasoning token budgets.

\begin{figure}[t]
    \centering
    \includegraphics[width=\linewidth]{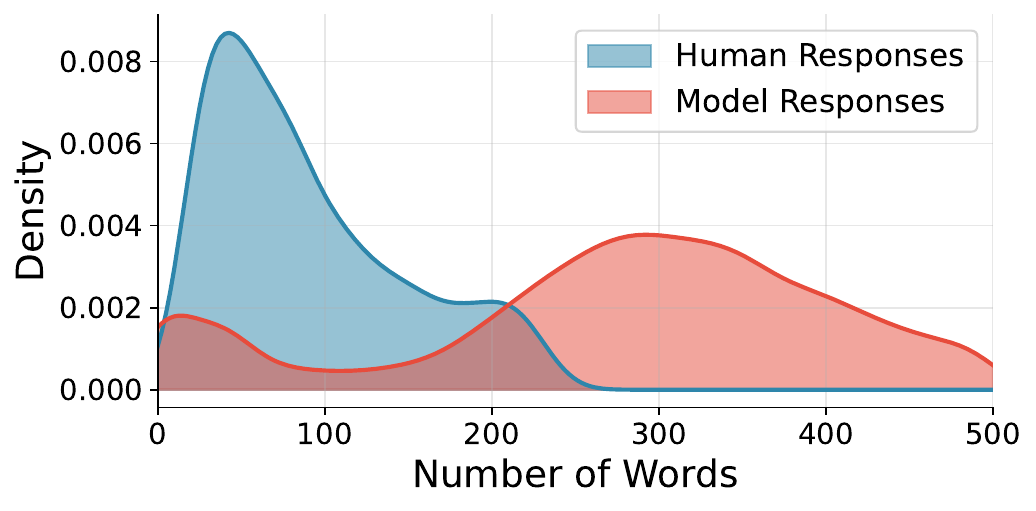}
    \vspace{-2.0em}
    \caption{\textbf{Models Talk More, Not Better.} Despite models generating $4$-$6\times$ more fatwas text than human, models do not achieve proportionally higher accuracy, indicating that verbosity serves as proxy for uncertainty rather than expertise.}
    \label{fig:verbosity}
    \vspace{-1.5em}
\end{figure}

\section{Results}

We analyze model behavior across tasks to understand the relationship between Arabic fluency and financial reasoning. Our findings highlight consistent patterns in performance, generalization, and failure modes across both recognition and generation settings.

\paragraph{Arabic Fluency ≠ Domain Reasoning: Event-Cause QA Exposes the Gap:} Arabic-centric pretraining provides strong foundations for Islamic jurisprudence tasks, but fails to transfer to financial reasoning (Accounting, Business). Domain-specific fine-tuning on \name{} closes this gap across all Arabic LLMs, with MCQ gains of +13.7\% (\textsc{Sahm-ALLAM-7B}), +5.8\% (\textsc{Sahm-Jais-8B}), and +5.2\% (SILMA-9B), enabling \textsc{Sahm-ALLAM-7B} to surpass GPT-5 on Accounting and Business and match 72B baselines. These improvements highlight the effectiveness of targeted domain adaptation in bridging reasoning gaps. Event-Cause QA emerges as the ``true IQ test'' for Arabic financial reasoning, exhibiting the widest performance spread ($1.89$--$9.84$), nearly the full scale.

Proprietary models cluster tightly at the top ($9.1$--$9.8$), followed by a sharp drop below $8.7$, exposing the limits of Arabic-centric models on causal financial reasoning. Language fluency does not imply domain reasoning: the task requires compositional causal inference that neither Arabic pretraining nor scale alone provides. Qualitative analysis (Figure~\ref{fig:error_qualitative}) reveals two dominant failure modes: ungrounded use of Islamic terminology (e.g., Gemma-3-27B fabricating \textit{\d{h}ad\={\i}th} evidence) and confusion between related financial concepts (e.g., Qwen2.5-72B miscomputing capitalizable interest).

\paragraph{The Recognition-Generation Gap:} A model that can identify correct Islamic rulings when presented as options should, in principle, generate coherent fatw\=as from scratch. Our results challenge this assumption. On Fatwa MCQ, Claude-Opus-4.5 and GPT-5 achieve $91.75\%$ and $90.75\%$ accuracy, respectively. However, their Fatwa QA scores drop to $8.81$ and $8.05$ out of $10$, a gap suggesting that recognition and generation tap fundamentally different competencies. Figure~\ref{fig:verbosity} illuminates one mechanism behind this gap. Human fatw\=as peak at approximately 50 words; model responses peak at 300 words, a $4$-$6\times$ inflation. Despite this verbosity, models do not achieve proportionally higher scores. We interpret this pattern as \emph{verbosity as uncertainty}: when models lack confident knowledge, they hedge with additional text rather than committing to precise rulings. This finding has practical implications: response length may signal answer confidence in Arabic financial QA systems, and evaluation protocols should distinguish between recognition and generation to avoid overestimating reliability.

\begin{table}[t]
\centering
\footnotesize
\resizebox{\linewidth}{!}{%
\begin{tabular}{lccc}
\toprule
\textbf{Model} & \textbf{ROUGE-1} & \textbf{ROUGE-2} & \textbf{ROUGE-L} \\
\midrule
\rowcolor{softgray}
\multicolumn{4}{c}{\textbf{Proprietary Models – Reasoning-Enhanced}} \\
\midrule
Claude-Opus-4.5            & 78.22 & 63.17 & 64.14 \\
GPT-5                      & 75.19 & 63.70 & 64.11 \\
Claude-Sonnet-4.5          & \textbf{79.86} & \textbf{64.98} &\textbf{ 65.13} \\
\midrule
\rowcolor{softgray}
\multicolumn{4}{c}{\textbf{Proprietary Models – General-Purpose}} \\
\midrule
Claude-Haiku-4.5           & \textbf{79.39} & 61.40 & 63.62 \\
GPT-4o-mini                & 77.79 & 62.90 & 64.08 \\
GPT-4o                     & 78.91 & \textbf{63.16} &\textbf{ 63.71} \\
Gemini-3-Flash             & 49.36 & 35.83 & 43.02 \\
Gemini-2.5-Flash           & 39.46 & 27.17 & 36.81 \\
\midrule
\rowcolor{softgray}
\multicolumn{4}{c}{\textbf{Open-source Models: $\geq$ 70B parameters}} \\
\midrule
Gemma-3-27B-IT             & \textbf{79.25} & \textbf{63.57} &\textbf{ 63.42} \\
Qwen2.5-72B-Instruct       & 40.52 & 29.50 & 34.04 \\
Meta-LLaMA-3.1-70B         & 39.64 & 31.40 & 32.65 \\
\midrule
\rowcolor{softgray}
\multicolumn{4}{c}{\textbf{Open-source Models: $<$ 70B parameters}} \\
\midrule
Qwen2.5-14B-Instruct       & 44.42 & 30.90 & 35.82 \\
Gemma-3-4B-IT              & \textbf{76.52} & \textbf{62.06} & \textbf{60.93} \\
Meta-LLaMA-3.1-8B          & 66.67 & 47.92 & 56.10 \\
Mixtral-8x7B-Instruct      & 32.71 & 13.07 & 23.78 \\
Qwen2.5-7B-Instruct        & 25.15 & 12.01 & 21.86 \\
\midrule
\rowcolor{softgray}
\multicolumn{4}{c}{\textbf{Arabic Models}} \\
\midrule
Jais-2-8B                    & \textbf{73.68} & \textbf{56.54} & \textbf{61.17} \\
Fanar-1-9B-Instruct          & 60.51 & 35.97 & 46.96 \\
ALLaM-7B-Instruct            & 35.97 & 22.61 & 28.24 \\
SILMA-9B-Instruct            & 27.92 & 16.66 & 25.99 \\
\bottomrule
\end{tabular}%
}
\caption{Extractive summarization performance on Arabic financial reports evaluated using ROUGE F1 (\%).}
\label{tab:extractive_rouge}
\end{table}\subsection{Extractive Summarization}
\label{sec:extractive_summarization}

Table~\ref{tab:extractive_rouge} reveals a striking inversion: Claude-Sonnet-4.5 achieves the highest ROUGE-1 (79.86), while Gemini-2.5-Flash a strong open-ended reasoner collapses to 39.46, underperforming even GPT-4o-mini (77.79). This exposes a fundamental tension: extractive summarization rewards \emph{verbatim selection}, not generative fluency. Consider a typical report: {\foreignlanguage{arabic}{``نجحت شركة بن غاطي للتطوير العقاري في طرح المزيد من الصكوك... بقيمة 300 مليون دولار أمريكي، ببورصة لندن وناسداك دبي''}} (Binghatti Development successfully issued additional sukuk... valued at \$300M, listed on the London Stock Exchange and Nasdaq Dubai). The gold summary must preserve the entity name, Islamic instrument (\textit{sukuk}), exact figure, and dual listing elements paraphrasing models systematically distort. Surprisingly, Gemma-3-4B-IT achieves 76.52 ROUGE-1, rivaling Claude-Opus-4.5 (78.22) with a fraction of the parameters, suggesting extraction benefits from constrained generation rather than extended reasoning. For Arabic-centric models, domain-specific tuning is decisive: SAHM-7B-Instruct attains 57.79 ROUGE-L, outperforming ALLaM-7B by \textbf{+29.55 points}, showing pretraining alone is insufficient.

\begin{figure}[t]
    \centering
    \includegraphics[width=\linewidth]{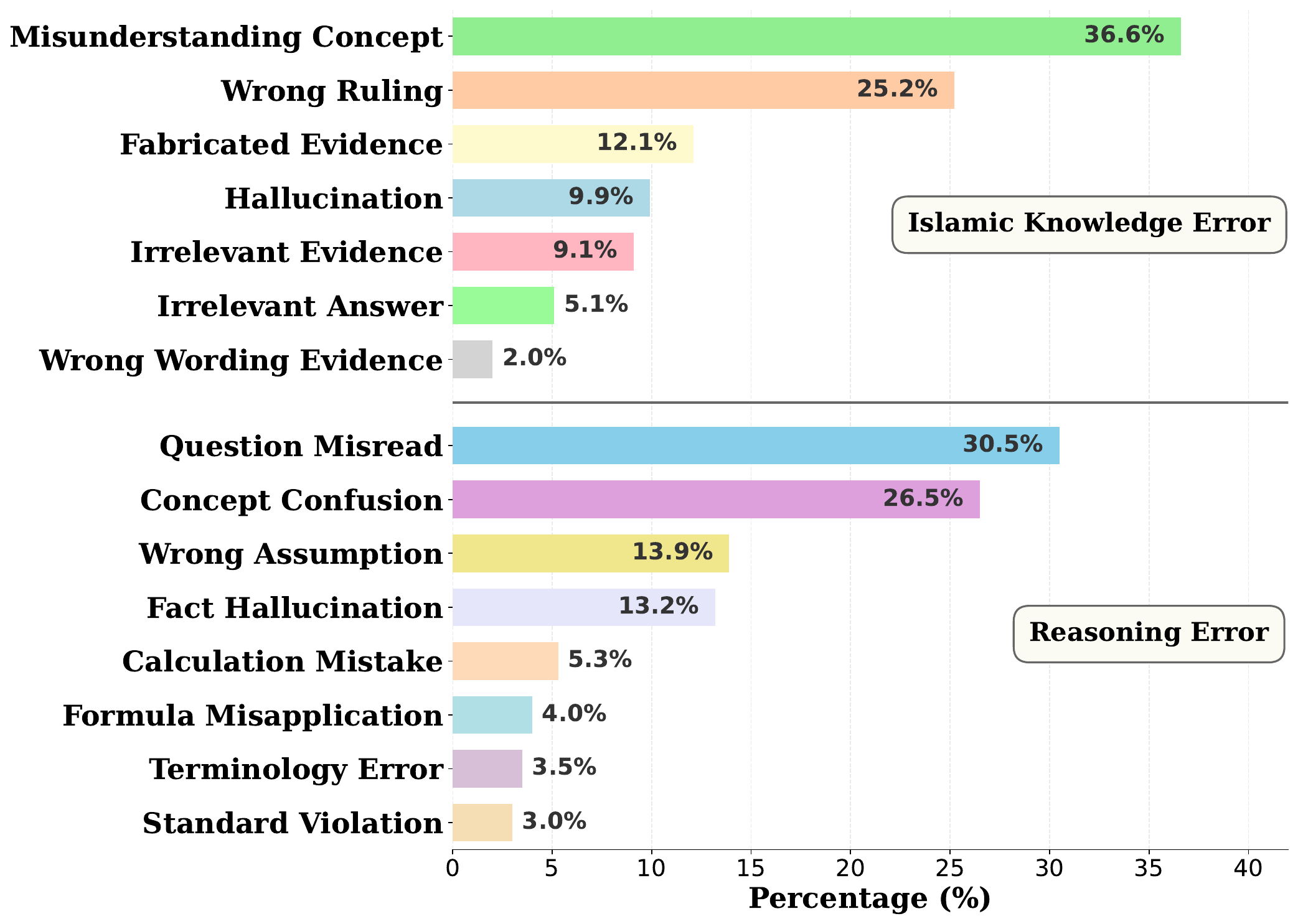}
    \vspace{-1.5em}
   \caption{Root cause distribution of model errors across Islamic knowledge and reasoning tasks.}
    \label{fig:root_cause_analysis}    
    \vspace{-1.5em}
\end{figure}

\subsection{Domain Adaptation Across Arabic LLMs}

We systematically evaluate domain adaptation across major Arabic-centric LLMs. Table~\ref{tab:finetuning_transfer} reveals three distinct adaptation profiles: (1) \textit{high-gain} bases like ALLAM yielding \textsc{Sahm-ALLAM-7B} with substantial improvement (+13.68\%), (2) \textit{stable-gain} bases like Jais-2 yielding \textsc{Sahm-Jais-8B} with consistent improvement across all MCQ metrics (+5.78\%, with notably strong gains in Sentiment +11.72\%), and (3) \textit{selective-gain} bases like SILMA that improve on some tasks (Sentiment +23.8\%) but regress on others (Accounting -7.8\%).

\begin{table*}[t]
\centering
\large
\resizebox{\linewidth}{!}{%
\begin{tabular}{lccccccccc}
\toprule
& \multicolumn{5}{c}{\textbf{MCQ (Accuracy \% $\uparrow$)}} & \multicolumn{4}{c}{\textbf{Open-Ended QA (Score 0–10 $\uparrow$)}} \\
\cmidrule(lr){2-6} \cmidrule(lr){7-10}
\textbf{Model} & \textbf{Accounting} & \textbf{Business} & \textbf{Fatw\=a} & \textbf{Sentiment} & \textbf{Mean} & \textbf{Event-Cause} & \textbf{Fatwa-QA} & \textbf{Islamic-Std} & \textbf{Mean} \\
\midrule
\rowcolor{softgray}
\multicolumn{10}{c}{\textbf{Base Models}} \\
\midrule
ALLAM-7B       & 44.91 & 68.31 & 74.40 & 58.75 & 61.59 & 6.89 & 4.94 & 4.22 & 5.35 \\
Jais-2-8B      & 35.33 & 60.30 & 66.10 & 46.25 & 52.00 & 4.69 & 2.51 & 4.24 & 3.81 \\
SILMA-9B       & 50.90 & 69.40 & 62.55 & 30.00 & 53.21 & 1.90 & 3.35 & 2.07 & 2.44 \\
\midrule
\rowcolor{softgray}
\multicolumn{10}{c}{\textbf{Fine-tuned Models}} \\
\midrule
\textbf{\textsc{Sahm-ALLAM-7B}}       & \textbf{71.40} {\small\textcolor{darkgreen}{(+26.5)}} & \textbf{93.99} {\small\textcolor{darkgreen}{(+25.7)}} & \textbf{74.45} {\small\textcolor{darkgreen}{(+0.1)}} & \textbf{61.25} {\small\textcolor{darkgreen}{(+2.5)}} & \textbf{75.27} {\small\textcolor{darkgreen}{(+13.7)}} & 6.79 {\small\textcolor{red}{(-0.1)}} & \textbf{6.48} {\small\textcolor{darkgreen}{(+1.5)}} & 4.12 {\small\textcolor{red}{(-0.1)}} & \textbf{5.80} {\small\textcolor{darkgreen}{(+0.5)}} \\
\textbf{\textsc{Sahm-Jais-8B}}        & 40.72 {\small\textcolor{darkgreen}{(+5.4)}} & 62.30 {\small\textcolor{darkgreen}{(+2.0)}} & 70.14 {\small\textcolor{darkgreen}{(+4.0)}} & 57.97 {\small\textcolor{darkgreen}{(+11.7)}} & 57.78 {\small\textcolor{darkgreen}{(+5.8)}} & \textbf{5.25} {\small\textcolor{darkgreen}{(+0.6)}} & 4.69 {\small\textcolor{darkgreen}{(+2.2)}} & \textbf{4.97} {\small\textcolor{darkgreen}{(+0.7)}} & 4.97 {\small\textcolor{darkgreen}{(+1.16)}} \\
SILMA-9B (fine-tuned)                 & 43.11 {\small\textcolor{red}{(-7.8)}} & 75.96 {\small\textcolor{darkgreen}{(+6.6)}} & 60.60 {\small\textcolor{red}{(-2.0)}} & 53.75 {\small\textcolor{darkgreen}{(+23.8)}} & 58.36 {\small\textcolor{darkgreen}{(+5.2)}} & 2.01 {\small\textcolor{darkgreen}{(+0.1)}} & 3.67 {\small\textcolor{darkgreen}{(+0.3)}} & 3.67 {\small\textcolor{darkgreen}{(+1.6)}} & 3.12 {\small\textcolor{darkgreen}{(+0.7)}} \\
\bottomrule
\end{tabular}%
}
\caption{Domain adaptation across Arabic LLMs. MCQ accuracy (\%) and Open-Ended QA scores (0--10) before and after fine-tuning on \name{}. \textbf{Bold model names} (\textsc{Sahm-ALLAM-7B}, \textsc{Sahm-Jais-8B}) denote the two released \name{}-family artifacts; SILMA-9B is included as a comparison case illustrating that adaptation outcomes depend on base-model properties.}
\label{tab:finetuning_transfer}
\end{table*}

\subsection{Error Analysis}
\label{sec:error_analysis}

To diagnose failure modes, we analyze 500 randomly sampled incorrect responses across all datasets, grouped by required competence: \textbf{Islamic Knowledge Errors} (Fatwa QA, Shari'ah Standards QA, Fatwa MCQ) and \textbf{Reasoning Errors} (Accounting, Business, Event-Cause QA); summarization errors are treated in \S\ref{sec:extractive_summarization} and sentiment via accuracy. The two annotators (see \S\ref{sec:sahm}) jointly adjudicated each error against the gold reference through a \textbf{consensus protocol}, additionally verifying cited religious evidence for Islamic tasks. Consensus over independent annotation with post-hoc IAA maximizes taxonomy coverage and ensures consistent categorization across heterogeneous errors requiring both jurisprudential and financial expertise.

\paragraph{Error Breakdown:} Figure~\ref{fig:root_cause_analysis} shows two dominant error types: \textit{Misunderstanding Concept} and \textit{Wrong Ruling}, accounting for 58.5\% of failures. Fabricated Evidence (11.4\%) and Hallucination (9.3\%) follow. Calculation errors are rare (0.3\%); models struggle not with arithmetic, but with selecting the correct computation.

\begin{figure}[!h]
    \centering
    \includegraphics[width=\linewidth]{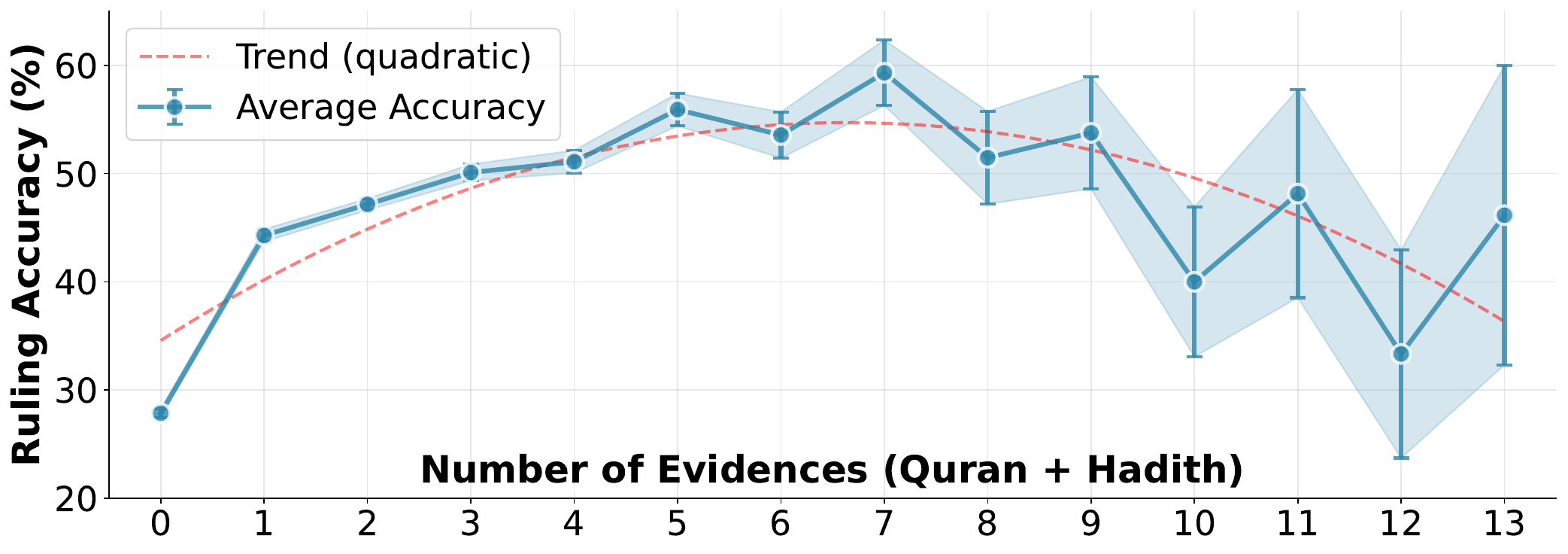}
    \caption{Effect of number of evidences from Hadith and Quran on Ruling Accuracy.}
    \label{fig:evidence_accuracy}
\end{figure}

\paragraph{Effect on Evidence Count on Accuracy.}
Figure~\ref{fig:evidence_accuracy} examines whether the presence of scriptural evidence (Qur'\=anic verses and \textit{\d{h}ad\={\i}th}) in reference answers correlates with model accuracy. We observe a logarithmic relationship: accuracy rises from 28\% with zero evidence to approximately 55\% with six or more citations. This pattern admits two interpretations. Optimistically, models may leverage textual evidence as grounding signals. Pessimistically, questions with more evidence may simply be easier or more frequently represented in training data. The increased variance at higher evidence counts (shaded region) suggests the relationship is not deterministic.

\section{Conclusion and Future Work}

We introduce \name{}, the first Arabic financial NLP benchmark integrating modern finance and Shari'ah-compliant reasoning across seven tasks. Evaluating 20 LLMs shows Arabic fluency does not imply financial reasoning, while fine-tuning on \name{} yields two models: \textsc{Sahm-ALLAM-7B} surpasses GPT-5 on Accounting and Business and matches 72B baselines, and \textsc{Sahm-Jais-8B} shows consistent gains across tasks, demonstrating that targeted domain adaptation outperforms scale. We release all resources to support trustworthy Arabic financial assistants.

Several directions extend this work. First, \name{} currently focuses on formal financial text; incorporating informal genres such as retail investor discourse, social media financial discussions, and dialectal Arabic would broaden coverage. Second, Arabic financial reports frequently contain tables, charts, and mixed-format documents; extending the benchmark to multimodal reasoning over structured financial data is a natural next step. Third, our evaluation assesses answer correctness but not evidence traceability; future metrics should explicitly verify cited Qur'anic verses, \textit{\d{h}ad\={\i}th} reports, and AAOIFI standard references. Fourth, cross-lingual transfer from English financial benchmarks to Arabic remains unexplored; investigating whether English financial reasoning capabilities transfer to Arabic could reduce data requirements. Finally, regional variation in Shari'ah interpretation across different supervisory bodies warrants task variants that evaluate model robustness to jurisdictional differences in Islamic finance rulings. 

More broadly, \name{} highlights the need for evaluation frameworks that move beyond surface-level language proficiency toward domain-grounded reasoning in high-stakes settings. As financial decision-making increasingly relies on LLMs, ensuring correctness, transparency, and alignment with regulatory and ethical standards becomes critical. Our findings also underscore the importance of culturally and legally informed benchmarks in shaping reliable AI systems for specialized domains. We hope this work motivates research on trustworthy Arabic financial AI, bridging language understanding, legal compliance, and real-world applicability, enabling safe and effective deployment across diverse financial and regulatory settings.

\section*{Limitations}
\noindent\textbf{Scope and coverage.} \name{} is built from curated, document-grounded sources and covers as much of the available public material as feasible; however, practical access and usage constraints on some online sources limit the extent to which additional genres can be incorporated at this time. As a result, while the benchmark provides strong provenance and reduces ambiguity, it does not yet cover all Arabic financial genres (e.g., informal retail-investor discourse) or fully capture regional and institutional variation in Arabic financial writing.

\noindent\textbf{Shari'ah-related content.} For Shari'ah-oriented questions, \name{} evaluates faithfulness to the referenced material and the reasoning constraints reflected in the provided sources; since interpretations may differ across jurisdictions and supervisory bodies, the benchmark is not intended to adjudicate between schools of thought, but rather to test source-grounded answering under the stated assumptions.

\noindent\textbf{Future evaluation directions.} As future work, we plan to develop evaluation metrics that explicitly assess (i) the existence and correctness of cited, source-verifiable evidence including traceable support from the underlying materials (e.g., fatwa text, and financial report statements) and, when answers cite religious evidence, the correctness of references such as Qur'anic verses, hadith reports, or named fiqh sources; and (ii) the accuracy of book/standard citations in model outputs (e.g., correct document title, section/article identifiers, and pointers that match the relevant source segment), enabling more direct measurement of citation faithfulness and evidence-groundedness.

\section*{Ethical Statement and Broad Impact}
\paragraph{Licensing.} We release \name{} under a dual license: (1) code and evaluation scripts under MIT License, and (2) annotation data under CC BY-NC 4.0, restricting commercial use while enabling academic research. Users must independently obtain source documents where applicable.
\paragraph{Availability.} The benchmark data, models, and evaluation scripts are publicly available at \url{https://huggingface.co/SahmBenchmark, https://github.com/mbzuai-nlp/SAHM}. 
\section*{Acknowledgments}
We acknowledge The Fin AI community for its research support, feedback, and collaborative environment that contributed to this work.

\bibliography{custom}


\appendix
\newpage
\appendix

\section{Islamic Finance Shari’ah Standards QA: Sources and Processing}
\label{sec:ocr-eval}

\paragraph{OCR Quality Evaluation.}
We developed a dedicated OCR quality evaluation tool to systematically assess recognition accuracy in Arabic legal–financial documents. The tool compares raw machine-extracted text against both the original scanned page and a manually corrected reference, enabling fine-grained verification of OCR fidelity at the page level. For each document, the system pairs a scanned page image (e.g., \texttt{page\_001.png}) with its corresponding OCR output (\texttt{page\_001.txt}) and presents them side by side: the original page image appears in the left panel, while the OCR-generated Arabic text is shown on the right. Annotators inspect these pairs to identify errors such as \foreignlanguage{arabic}{نص مفقود} (missing text), \foreignlanguage{arabic}{أحرف غير صحيحة} (incorrect characters), \foreignlanguage{arabic}{ترتيب كلمات خاطئ} (incorrect word order), and \foreignlanguage{arabic}{فقدان التنسيق} (formatting loss). When needed, they correct the OCR output using an editable field while monitoring a live similarity score reflecting the edit distance between the corrected and original text.

In addition to corrections, annotators label common OCR failure modes, including distorted symbols (\foreignlanguage{arabic}{رموز خاصة مشوهة}), punctuation errors (\foreignlanguage{arabic}{أخطاء علامات الترقيم}), and inaccurate numerals (\foreignlanguage{arabic}{أرقام غير دقيقة}), and may add comments on recurring issues such as confusion between similar characters (e.g., \foreignlanguage{arabic}{ب} vs. \foreignlanguage{arabic}{ن}) or misinterpreted \foreignlanguage{arabic}{التشكيل}. The system computes a quality score via character-level edit distance, mapping similarity to four categories: \textit{Excellent} ($\geq$95\%), \textit{Good} (80--95\%), \textit{Partial} (50--80\%), and \textit{Poor} ($<$50\%). All steps are logged as structured JSON records (text, scores, error types, comments, timestamps), ensuring reproducibility and auditability.

\begin{figure*}[t]
  \centering
  \includegraphics[width=0.8\textwidth, trim=0 400 0 0, clip]{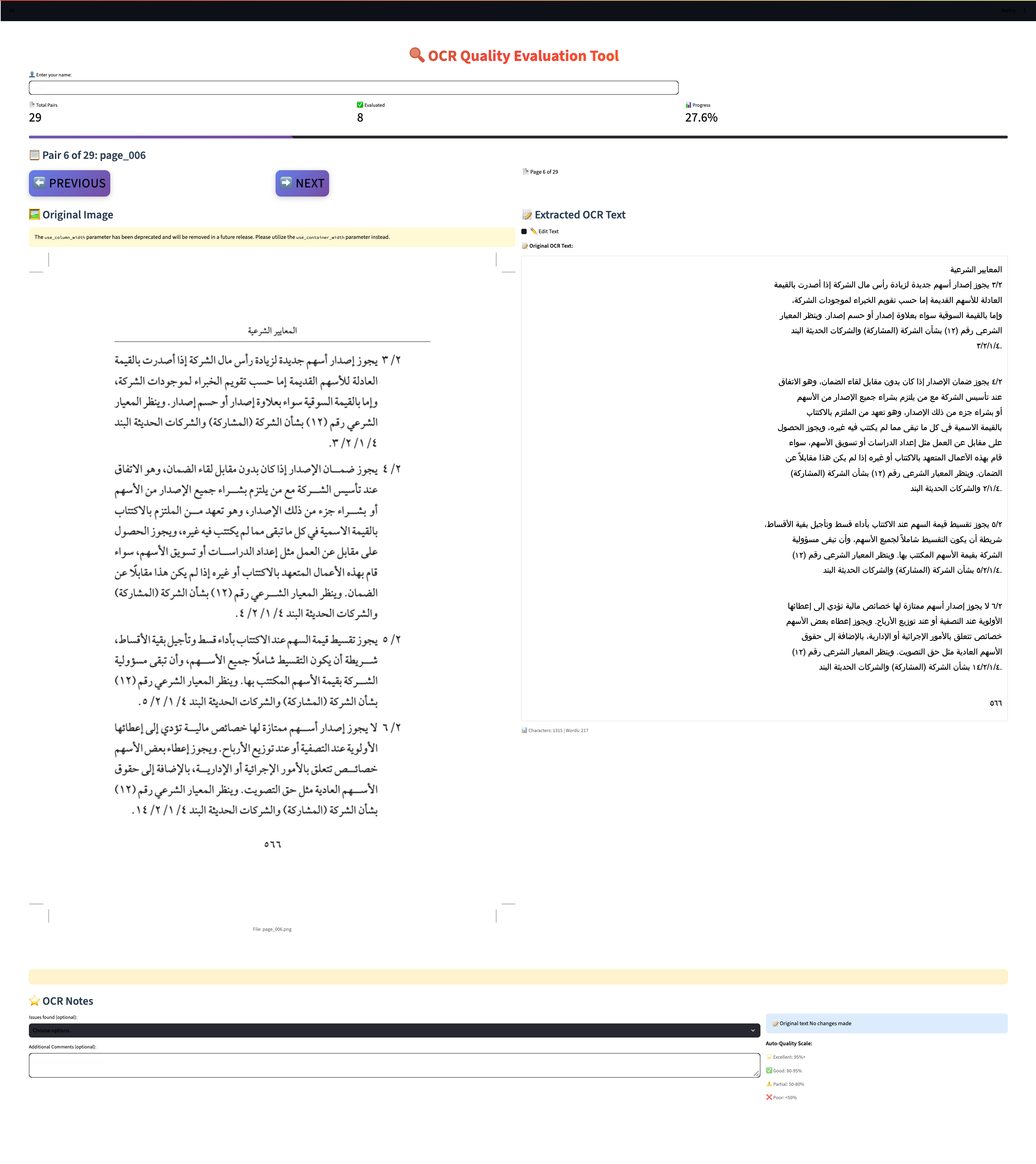}
  \caption{OCR quality evaluation interface for the Shari’ah Standards QA dataset. The tool displays each scanned page from the AAOIFI Shari’ah Standards (left) alongside the OCR-extracted Arabic text (right) to support manual quality verification. Annotators compare the original page with the extracted text, flag recognition errors in diacritics, numerals, and domain-specific terminology, and add corrective notes (bottom). A progress bar tracks annotation completion and overall OCR accuracy.}
  \label{fig:ocr_eval_interface}
\end{figure*}

Beyond per-page inspection, the pipeline enables aggregate OCR analysis across document collections, identifying systematic errors, benchmarking quality across diverse Arabic sources, and informing downstream normalization and model refinement. Overall, this human-in-the-loop approach ensures that OCR text used in Arabic financial NLP benchmarks and training is accurate and free from errors that could affect \foreignlanguage{arabic}{الاستدلال الشرعي} (jurisprudential reasoning) or \foreignlanguage{arabic}{التحليل المالي} (financial analysis). Figure~\ref{fig:ocr_eval_interface} illustrates the annotation interface.

\paragraph{Prompt Design:} To ensure high-quality OCR extraction, we design a constrained prompt (Figure~\ref{fig:ocr_prompt}) that enforces verbatim transcription, preservation of diacritics and formatting, and strict exclusion of non-textual artifacts. These constraints are critical for maintaining fidelity in Arabic legal documents, where minor textual variations can alter meaning. Similarly, we construct a controlled prompt for question--answer generation (Figure~\ref{fig:qa_prompt}) that restricts outputs to the explicit content of the Shari’ah standards. This design prevents hallucination, preserves juridical precision, and ensures that generated QA pairs remain faithful to the source text.

\begin{figure}[!h]
\centering
\begin{tcolorbox}[title={Prompt for Arabic OCR Text Extraction}, width=\linewidth]
\small

\textbf{Task.}
You are an expert Arabic OCR system specialized in legal and financial documents. Given a scanned page image from an official Islamic finance standard, your task is to extract the text \emph{verbatim} in Arabic with maximum fidelity to the original source.

\vspace{2pt}
\textbf{Extraction guidelines.}
\begin{itemize}
  \item Preserve the original wording exactly; do \emph{not} paraphrase, summarize, or infer missing content.
  \item Preserve diacritics (\foreignlanguage{arabic}{التشكيل}) whenever present in the source.
  \item Preserve all numerals exactly as written; do not normalize or convert number formats.
  \item Preserve punctuation, headings, lists, and paragraph boundaries as faithfully as possible.
  \item Do not correct perceived grammatical, typographical, or stylistic issues.
  \item If text is unclear or partially illegible, extract the most faithful representation without guessing.
\end{itemize}

\vspace{2pt}
\textbf{What to ignore.}
\begin{itemize}
  \item Page numbers, running headers, footers, or decorative elements not part of the main content.
  \item Marginal artifacts or scanning noise that do not belong to the text.
\end{itemize}

\vspace{2pt}
\textbf{Critical rule.}
Do \emph{not} add explanations, comments, translations, or annotations. Output Arabic text only.

\vspace{2pt}
\textbf{Input.}
A single scanned page image from the AAOIFI Shari’ah Standards.

\vspace{2pt}
\textbf{Output format.}
Return the extracted Arabic text as plain UTF-8 text, preserving line breaks and paragraph structure.

\end{tcolorbox}
\caption{Prompt for Arabic OCR text extraction with strict verbatim fidelity and formatting preservation.}
\label{fig:ocr_prompt}
\end{figure}

\section{Islamic Fatwa Dataset: Sources and Processing}
\label{sec:fatwa_prompt}
\begin{figure}[!h]
\centering
\begin{tcolorbox}[title={Prompt for Shari’ah Standards Question--Answer Generation}, width=\linewidth]
\small

\textbf{Task.}
You are an assistant supporting the creation of evaluation data for Islamic finance. Given a verified excerpt from an official Shari’ah standard, your task is to draft candidate Arabic question--answer pairs that reflect the \emph{explicit ruling stated in the text}.

\vspace{2pt}
\textbf{Question generation.}
\begin{itemize}
  \item Formulate a clear, focused question that asks about the ruling, condition, or permissibility described in the excerpt.
  \item Do not introduce hypothetical scenarios or facts not present in the source text.
  \item Ensure the question can be answered directly and completely from the provided excerpt.
\end{itemize}

\vspace{2pt}
\textbf{Answer generation.}
\begin{itemize}
  \item Base the answer strictly on the given excerpt; do not add external knowledge.
  \item Preserve the original legal meaning, mandatory conditions, and stated exceptions.
  \item Do not simplify, reinterpret, or generalize the ruling beyond what the text explicitly states.
  \item Use formal Arabic consistent with fiqh al-muʿāmalāt terminology.
\end{itemize}

\vspace{2pt}
\textbf{Restrictions.}
\begin{itemize}
  \item Do not issue personal opinions or normative judgments.
  \item Do not cite sources outside the provided text.
  \item Do not omit conditions, constraints, or qualifiers that affect the ruling.
\end{itemize}

\vspace{2pt}
\textbf{Input template.}
\texttt{STANDARD\_EXCERPT: \{arabic\_text\}}

\vspace{2pt}
\textbf{Output format.}
\begin{verbatim}
{
  "question": "...",
  "answer": "..."
}
\end{verbatim}

\end{tcolorbox}
\caption{Prompt for generating Arabic QA pairs from Shari’ah standard excerpts with strict fidelity to explicit rulings.}
\label{fig:qa_prompt}
\end{figure}

\begin{figure*}[t]
  \centering
  \includegraphics[width=0.8\textwidth, trim=0 400 0 0, clip]{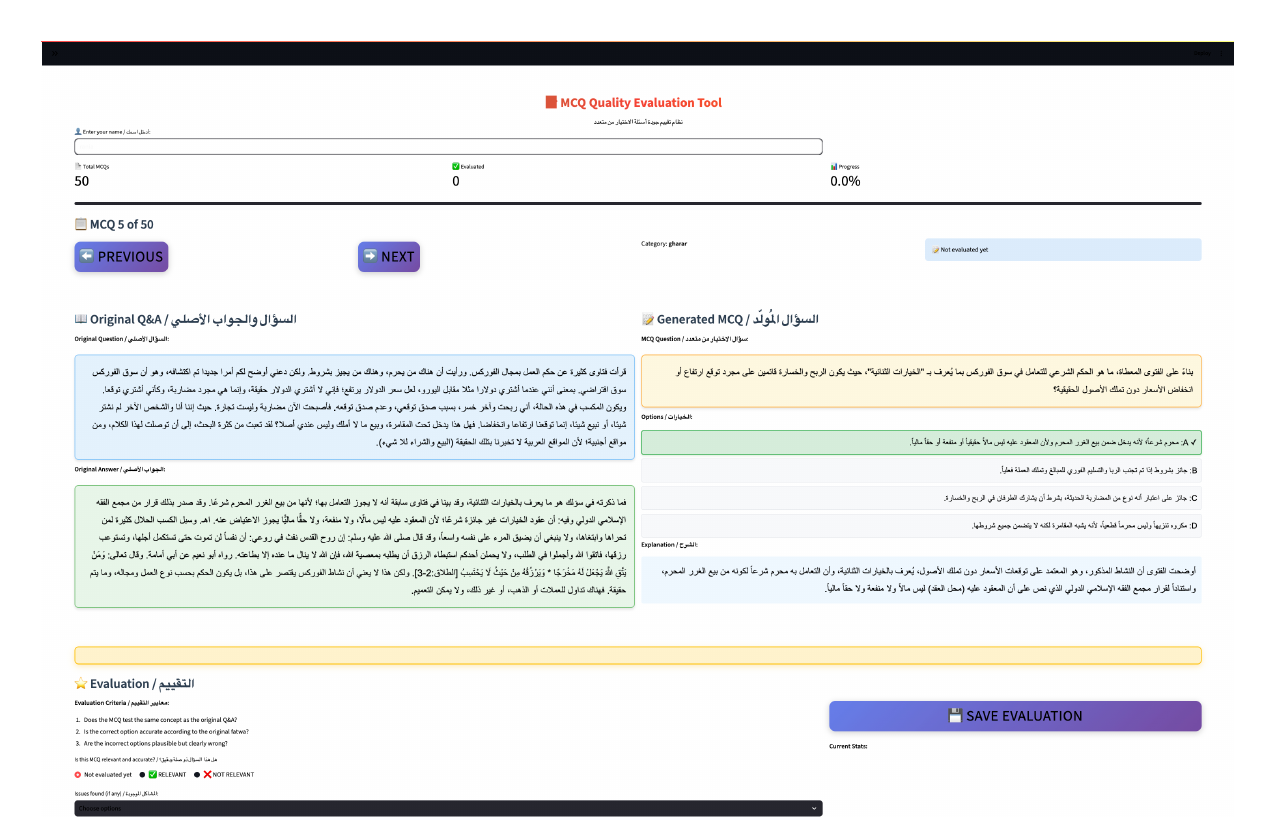}
  \caption{Custom annotation interface used to validate automatically generated multiple-choice questions (MCQs) for the Islamic Finance Fatwa Q\&A dataset. The interface displays each original question–answer pair on the left and the corresponding AI-generated MCQ on the right, including the question, answer options, and the automatically selected correct choice. Annotators review conceptual alignment between the MCQ and the original fatwā, verify the correctness and terminology of the marked answer, and assess the plausibility and pedagogical value of distractors. The bottom panel provides structured evaluation criteria and issue tagging to ensure consistent, high-quality validation.}
  \label{fig:mcq_annotation_interface}
\end{figure*}

\label{appendix:fatwa}
\begin{table*}[!h]
\centering 
\small
\setlength{\tabcolsep}{6pt} 
\renewcommand{\arraystretch}{1.2} 
\begin{tabular}{lll} \toprule \textbf{Website} & \textbf{Link} & \textbf{Country} \\ \midrule Dar Al Ifta in Saudi Arabia & \href{https://www.alifta.gov.sa/}{https://www.alifta.gov.sa/} & Saudi Arabia \\ Dar Al Ifta in Egypt & \href{https://www.dar-alifta.org}{https://www.dar-alifta.org} & Egypt \\ Dar Al Ifta in Jordan & \href{https://aliftaa.jo}{https://aliftaa.jo} & Jordan \\ Al Shaikh Abdual Aziz Ibn Baz & \href{https://binbaz.org.sa}{https://binbaz.org.sa} & Saudi Arabia \\ Al Shaikh Mohammad Ibn Othaimin & \href{https://binothaimeen.net/site}{https://binothaimeen.net/site} & Saudi Arabia \\ Al Shaikh Abdual Aziz Al Ashaikh & \href{https://www.mufti.af.org.sa}{https://www.mufti.af.org.sa} & Saudi Arabia \\ Al Shaikh Saleh Al Fwzan & \href{https://www.alfawzan.af.org.sa}{https://www.alfawzan.af.org.sa} & Saudi Arabia \\ Al Shaikh Saleh Bin Humaid & \href{https://www.ibnhomaid.af.org.sa/}{https://www.ibnhomaid.af.org.sa/} & Saudi Arabia \\ Al Shaikh Abdullah Al Manee & \href{https://al-manee.com}{https://al-manee.com} & Saudi Arabia \\ IslamWeb & \href{https://www.islamweb.com}{https://www.islamweb.com} & Qatar \\ FatwaPedia & \href{https://fatwapedia.com}{https://fatwapedia.com} & Saudi Arabia \\ IslamQA & \href{https://islamqa.info}{https://islamqa.info} & Syria \\ IslamOnline & \href{https://islamonline.net}{https://islamonline.net} & Qatar \\ \bottomrule \end{tabular} \caption{Primary online fatwā archives used for collecting Islamic financial question–answer pairs. These official and widely recognized sites span seven Arab countries, providing diverse juristic opinions and real-world financial scenarios. The URLs shown correspond to the original Arabic portals from which data was programmatically scraped and later cleaned for inclusion in the dataset.} \label{tab:fatwa_websites} 
\end{table*}

The purpose of this evaluation is to determine whether an AI-generated multiple-choice question (MCQ) accurately tests the same Islamic jurisprudence concept as the original \foreignlanguage{arabic}{فتوى} Q\&A pair. The goal is to maintain both pedagogical soundness and factual correctness. A well-formed MCQ must remain conceptually aligned with the original ruling (\foreignlanguage{arabic}{الحكم الشرعي}), preserve the main \foreignlanguage{arabic}{مفهوم فقهي} without distortion, and use appropriate \foreignlanguage{arabic}{مصطلحات فقهية} to reflect the opinion of the original scholar (\foreignlanguage{arabic}{المُفتي}). Evaluators must ensure that the question targets the central legal issue and does not introduce unrelated details or alter the scenario in a way that changes the ruling. This evaluation is conducted through a structured annotation dashboard (Figure~\ref{fig:mcq_annotation_interface}) that presents the original fatwā alongside the generated MCQ for systematic validation. The fatwā Q\&A pairs used in this evaluation are collected from a diverse set of authoritative online sources (Table~\ref{tab:fatwa_websites}). 

\begin{figure}[!h]
\centering
\begin{tcolorbox}[title={Prompt for Fatwā Q\&A Normalization}, width=\linewidth]
\small
\textbf{Task.}
You are an expert Arabic copy-editor specializing in Islamic finance Q\&A. Given a \texttt{QUESTION} and an \texttt{ORIGINAL ANSWER}, your goal is to produce a concise, self-contained question and answer pair in Arabic by removing only non-essential elements \emph{without paraphrasing or changing the juristic intent}. Do \emph{not} summarize or rephrase; keep the original wording as much as possible.

\vspace{2pt}
\textbf{1. Referral flag.}
Before editing, set \texttt{IS\_MAINLY\_REFERRAL}:
\begin{itemize}
  \item \texttt{"YES"} if the answer mainly redirects to another fatwā, link, or reference and does not provide a substantive independent ruling.
  \item \texttt{"NO"} otherwise.
\end{itemize}

\vspace{2pt}
\textbf{2. Clean the question.}
Edit minimally while preserving wording and fiqh intent:
\begin{itemize}[noitemsep, topsep=2pt, leftmargin=*]
  \item Remove greetings, honorifics, and personal appeals (e.g., \foreignlanguage{arabic}{سماحة الشيخ}، \foreignlanguage{arabic}{سلمه الله}، \foreignlanguage{arabic}{السلام عليكم}).
  \item Remove formal closings (e.g., \foreignlanguage{arabic}{أرجو منكم التكرم}، \foreignlanguage{arabic}{وجزاكم الله خيراً}).
  \item Remove the scholar’s name if it is only a form of address; keep it only if the question explicitly seeks that scholar’s specific fatwā or opinion.
  \item Ensure the final question reads as a natural, standalone query.
\end{itemize}

\vspace{2pt}
\textbf{3. Clean the answer.}
Edit minimally while preserving wording and reasoning:
\begin{itemize}[noitemsep, topsep=2pt, leftmargin=*]
  \item Remove formal openings and closings so the answer starts with substantive content.
  \item Remove all fatwā numbers, hyperlinks, and navigational phrases, editing surrounding text just enough to remain grammatical.
  \item Convert Arabic-Indic numerals to Western numerals.
  \item Remove purely formulaic closings such as \foreignlanguage{arabic}{وفقكم الله} and \foreignlanguage{arabic}{والله أعلم} when they are not part of practical advice.
  \item Always preserve Qurʾānic verses and sūrah references, ḥadīth attributions, and citations of scholars and their opinions.
\end{itemize}

\vspace{2pt}
\textbf{Global rule.}
Always delete \emph{all} fatwa numbers from the cleaned question and cleaned answer.

\vspace{2pt}
\textbf{Input template.}
\texttt{TITLE: \{title\}}\\
\texttt{QUESTION: \{question\}}\\
\texttt{ORIGINAL ANSWER: \{answer\}}

\vspace{2pt}
\textbf{Output format.}
\begin{verbatim}
{
  "IS_MAINLY_REFERRAL": "YES" or "NO",
  "cleaned_question": "...",
  "cleaned_answer": "..."
}
\end{verbatim}
\end{tcolorbox}
\caption{Prompt for Arabic fatwā Q\&A normalization with minimal editing and preservation of juristic intent.}
\label{fig:fatwa_prompt}
\end{figure}

For an MCQ to be marked as \foreignlanguage{arabic}{ملائم} (RELEVANT), it must meet four criteria. First, conceptual alignment (\foreignlanguage{arabic}{المواءمة المفاهيمية}): the question should test the same core ruling as the source fatwa and remain faithful to its reasoning and conditions. Second, answer accuracy (\foreignlanguage{arabic}{دقة الإجابة الصحيحة}): the correct option must match the original answer, be contradiction-free, and use precise legal terms. Third, distractor quality (\foreignlanguage{arabic}{جودة الخيارات الخاطئة}): incorrect options should be plausible yet clearly wrong, reflecting common misunderstandings. Finally, question clarity (\foreignlanguage{arabic}{وضوح السؤال}) the MCQ must be clearly phrased, grammatically correct in \foreignlanguage{arabic}{العربية}, and provide enough context to be answerable without referencing the original text.

Conversely, an MCQ should be marked as \foreignlanguage{arabic}{غير ملائم} (NOT RELEVANT) if it fails any major requirement. Conceptual misalignment occurs when the question tests a different topic, oversimplifies a complex juristic issue, or changes critical context such as conditions (\foreignlanguage{arabic}{شروط}) or scenarios. Incorrect answer issues include a keyed option that contradicts the fatwa, multiple potentially correct answers, or misleading explanations. Poor distractor quality arises when wrong options are obviously incorrect, factually wrong about \foreignlanguage{arabic}{الإسلام}, or too ambiguous. Technical problems include grammar errors that affect meaning, vague or incomplete questions, or improper mixing of different \foreignlanguage{arabic}{مذاهب} in a way that confuses the intended ruling.

The evaluation process follows a clear four-step workflow. First, read the original Q\&A carefully, identify the primary \foreignlanguage{arabic}{حكم}, any \foreignlanguage{arabic}{شروط} or exceptions, and the supporting evidence such as Qur’anic verses or \foreignlanguage{arabic}{حديث}. Second, analyze the generated MCQ to check conceptual consistency, correct answer faithfulness, and plausibility of distractors. 

Third, look for red flags such as contradictions, oversimplification, missing qualifiers, or scenario changes. Finally, make a decision: label the MCQ as \foreignlanguage{arabic}{ملائم} if it meets all core criteria (minor language or formatting issues may be tolerated) or as \foreignlanguage{arabic}{غير ملائم} if any critical issue is present. This structured approach ensures that evaluation is consistent, transparent, and preserves the integrity of Islamic legal reasoning in AI-generated questions. 

\paragraph{Normalization Prompt.}
To standardize fatwā question--answer pairs, we design a constrained prompt (Figure~\ref{fig:fatwa_prompt}) that removes non-essential elements such as greetings and formatting artifacts while preserving the original wording and legal intent. This ensures consistency across examples without altering the underlying \foreignlanguage{arabic}{حكم شرعي}.

\section{Evaluated Models}
\label{sec:model_discussion}

This appendix briefly documents the rationale behind the selection of models evaluated in Table~\ref{modelperformance}, with model specifications summarized separately in Table~\ref{tab:models}. The goal is not comparative analysis, but transparency regarding model coverage across language focus, scale, and accessibility.

\begin{table*}[t]
\centering
\scriptsize
\setlength{\tabcolsep}{6pt}
\renewcommand{\arraystretch}{1.2}
\begin{tabular}{llll}
\toprule
\textbf{Model} & \textbf{Organization} & \textbf{Size} & \textbf{Source / Notes} \\
\midrule
\multicolumn{4}{l}{\textbf{Arabic-Focused Models}} \\
\midrule
ALLAM-7B-Instruct & SDAIA / ALLaM-AI & 7B & \cite{bari2024allamlargelanguagemodels} \\
Fanar-1-9B-Instruct & QCRI & 9B & \cite{fanar} \\
SILMA-9B-Instruct & SILMA AI & 9B & \cite{silma-9b-2024} \\
\midrule
\multicolumn{4}{l}{\textbf{Strong Multilingual / General Open-Source Models}} \\
\midrule
Qwen2.5-72B-Instruct & Alibaba & 72B & \cite{qwen25} \\
LLaMA-3.1-70B-Instruct & Meta & 70B & \cite{llama3} \\
Qwen2.5-14B-Instruct & Alibaba & 14B & \cite{qwen25} \\
Qwen2.5-7B-Instruct & Alibaba & 7B & \cite{qwen25} \\
Gemma-2-9B-IT & Google & 9B & \cite{gemma2} \\
Gemma-3-27B-IT & Google & 27B & \cite{gemma3} \\
Gemma-3-4B-IT & Google & 4B & \cite{gemma3} \\
LLaMA-3.1-8B-Instruct & Meta & 8B & \cite{llama3} \\
Mixtral-8x7B-Instruct & Mistral AI & 8$\times$7B & \cite{jiang2024mixtralexperts} \\
\midrule
\multicolumn{4}{l}{\textbf{Proprietary Models (Upper-Bound References)}} \\
\midrule
GPT-5 & OpenAI & -- & \cite{gpt5} (API) \\
GPT-4o & OpenAI & -- & \cite{gpt4o} (API) \\
GPT-4o-mini & OpenAI & -- & \cite{gpt4o} (API) \\
Claude Opus 4.5 & Anthropic & -- & \cite{claude45opus} (API) \\
Claude Sonnet 4.5 & Anthropic & -- & \cite{claude45sonnet} (API) \\
Claude Haiku 4.5 & Anthropic & -- & \cite{claude45haiku} (API) \\
Gemini-3-Flash (preview) & Google DeepMind & -- & \cite{gemini3flash} (API) \\
\bottomrule
\end{tabular}
\caption{Models evaluated in this study, grouped into Arabic-focused models, strong multilingual open-source baselines, and proprietary frontier models used as upper-bound references.}
\label{tab:models}
\end{table*}




\paragraph{Arabic-Focused Models.}
We include ALLAM-7B-Instruct~\cite{bari2024allamlargelanguagemodels}, Fanar-1-9B-Instruct~\cite{fanar}, and SILMA-9B-Instruct~\cite{silma-9b-2024} as representative publicly available Arabic-adapted instruction-tuned models. These systems span different base architectures and training strategies, capturing the diversity of current Arabic-centric efforts.

\paragraph{Open-Source Multilingual Models.}
We evaluate Qwen2.5~\cite{qwen25}, LLaMA-3.1~\cite{llama3}, Gemma-2/3~\cite{gemma2,gemma3}, and Mixtral-8x7B-Instruct~\cite{jiang2024mixtralexperts} as strong open-weight baselines across scales. These widely used models provide reference points for general multilingual performance on Arabic financial and jurisprudential tasks.

\paragraph{Proprietary Models.}
GPT-5~\cite{gpt5}, GPT-4o~\cite{gpt4o}, Claude-4.5 variants~\cite{claude45opus,claude45sonnet,claude45haiku}, and Gemini-3-Flash~\cite{gemini3flash} serve as closed-source upper-bound references, contextualizing open and Arabic-focused models against frontier systems without implying direct comparability.

\section{Business and Accounting Exam Extraction Prompts}
\label{app:exam_prompts}


Business and accounting exams in Arabic exhibit heterogeneous layouts, ranging from narrative exercise-based formats to tabular true/false questions, often with inconsistent formatting and varying levels of structural clarity across documents. To reliably extract structured MCQs from these sources, we use two task-specific prompts tailored to the dominant document formats observed in the collected exams: an exercise-based extraction prompt (Figure~\ref{fig:accounting_prompt}) and a table-oriented extraction prompt (Figure~\ref{fig:business_prompt}), enabling robust handling of both free-form and semi-structured exam content across diverse real-world settings.

\section{Financial Sentiment Annotation Guidelines}
\label{app:sentiment_guidelines}

\begin{figure*}[t]
  \centering
  \includegraphics[width=1.0\textwidth, trim=0 400 0 0, clip]{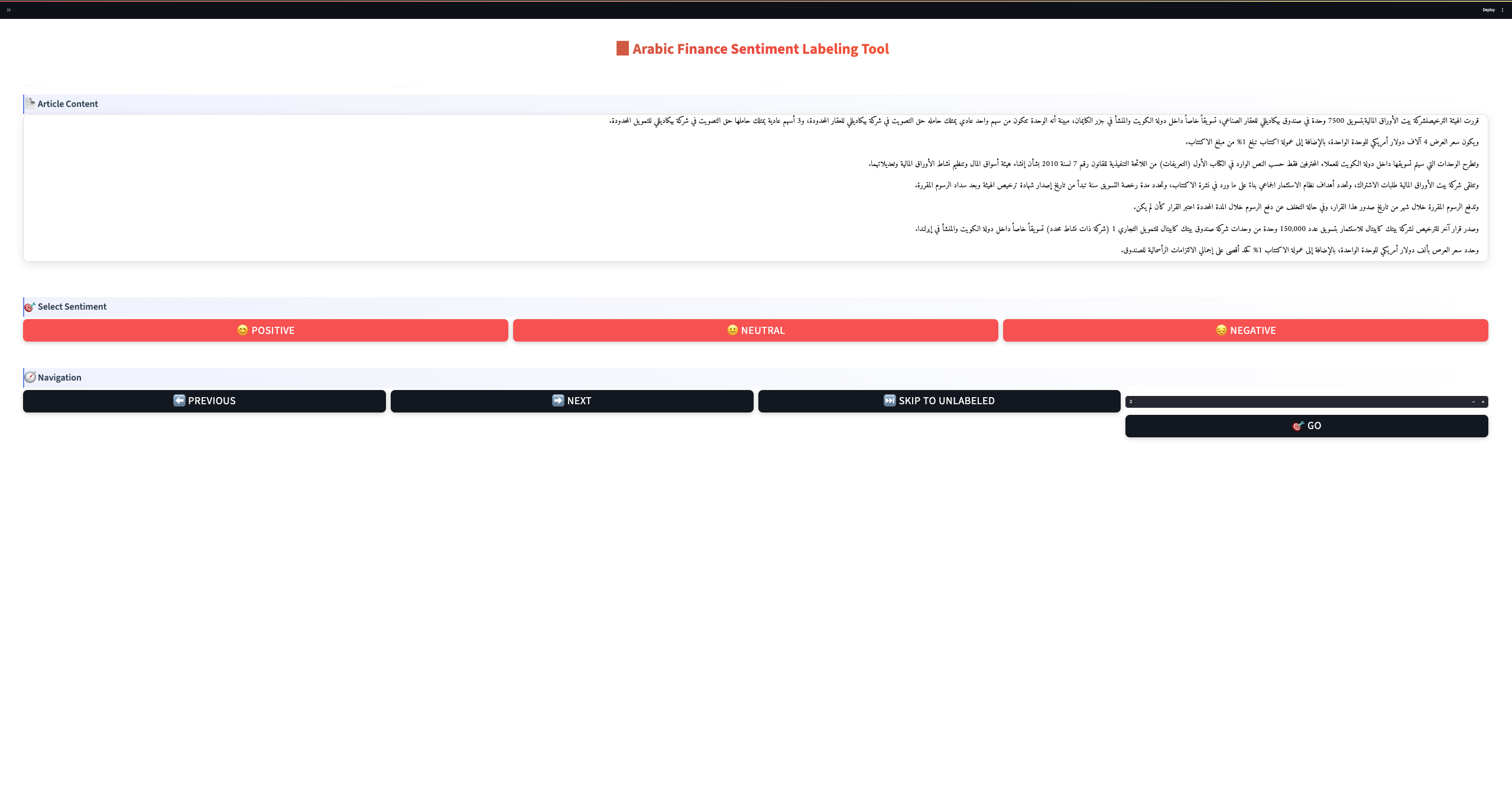}
  \caption{Custom annotation platform used to label Arabic financial reports for sentiment analysis. Annotators reviewed full reports, assigned sentiment classes, and flagged ambiguous cases for expert adjudication.}
  \label{fig:sentiment_annotation}
\end{figure*}


We annotate Arabic financial reports using a document-level sentiment scheme designed to reflect overall market impact rather than sentence-level polarity. Annotation follows a structured human-in-the-loop workflow supported by a custom web-based interface (Figure~\ref{fig:sentiment_annotation}), with clear, well-defined decision rules (Figure~\ref{fig:sentiment_guidelines}) to ensure consistency across Islamic and conventional financial reporting contexts.

\begin{figure}[!h]
\centering
\begin{tcolorbox}[title={Document-Level Financial Sentiment Annotation Guidelines}, width=\linewidth]
\small

\textbf{Core principle.}  
Assign a single sentiment label based on the \emph{overall dominant sentiment of the entire report}, not on individual sentences or isolated phrases.

\vspace{2pt}
\textbf{Annotation procedure.}
\begin{itemize}[noitemsep, topsep=2pt, leftmargin=*]
  \item Read the complete document before assigning any label.
  \item Identify the main financial outcome, thesis, and conclusion.
  \item Give greater weight to headlines, executive summaries, and concluding sections than to supporting details.
\end{itemize}

\vspace{2pt}
\textbf{Handling mixed sentiment.}
\begin{itemize}
  \item \textbf{Dominant sentiment rule}: assign \texttt{Positive} or \texttt{Negative} if one polarity accounts for more than 60\% of the salient content.
  \item \textbf{Neutral default}: assign \texttt{Neutral} when positive and negative signals are balanced  or when the report is primarily factual.
\end{itemize}

\vspace{2pt}
\textbf{Decision criteria.}
\begin{itemize}[noitemsep, topsep=2pt, leftmargin=*]
  \item \textbf{Positive}: growth announcements, profit increases, successful expansions, or favorable forecasts.
  \item \textbf{Negative}: losses, declining performance, regulatory issues, or adverse outlooks.
  \item \textbf{Neutral}: factual reporting, balanced analysis, or informational updates without a clear directional impact.
\end{itemize}

\vspace{2pt}
\textbf{Quality control.}  
Annotators label reports independently, resolve disagreements during a calibration phase, and refine shared decision criteria. A third domain expert adjudicates remaining conflicts. Each report receives exactly one final sentiment label.

\end{tcolorbox}
\caption{Guidelines for document-level sentiment annotation of Arabic financial reports.}
\label{fig:sentiment_guidelines}
\end{figure}

\section{Arabic Finance Extractive Summarization Annotation Guidelines}
\label{sec:report-extractive-summarization}

\begin{figure*}[t]
  \centering
  \includegraphics[width=1.0\textwidth, trim=0 400 0 0, clip]{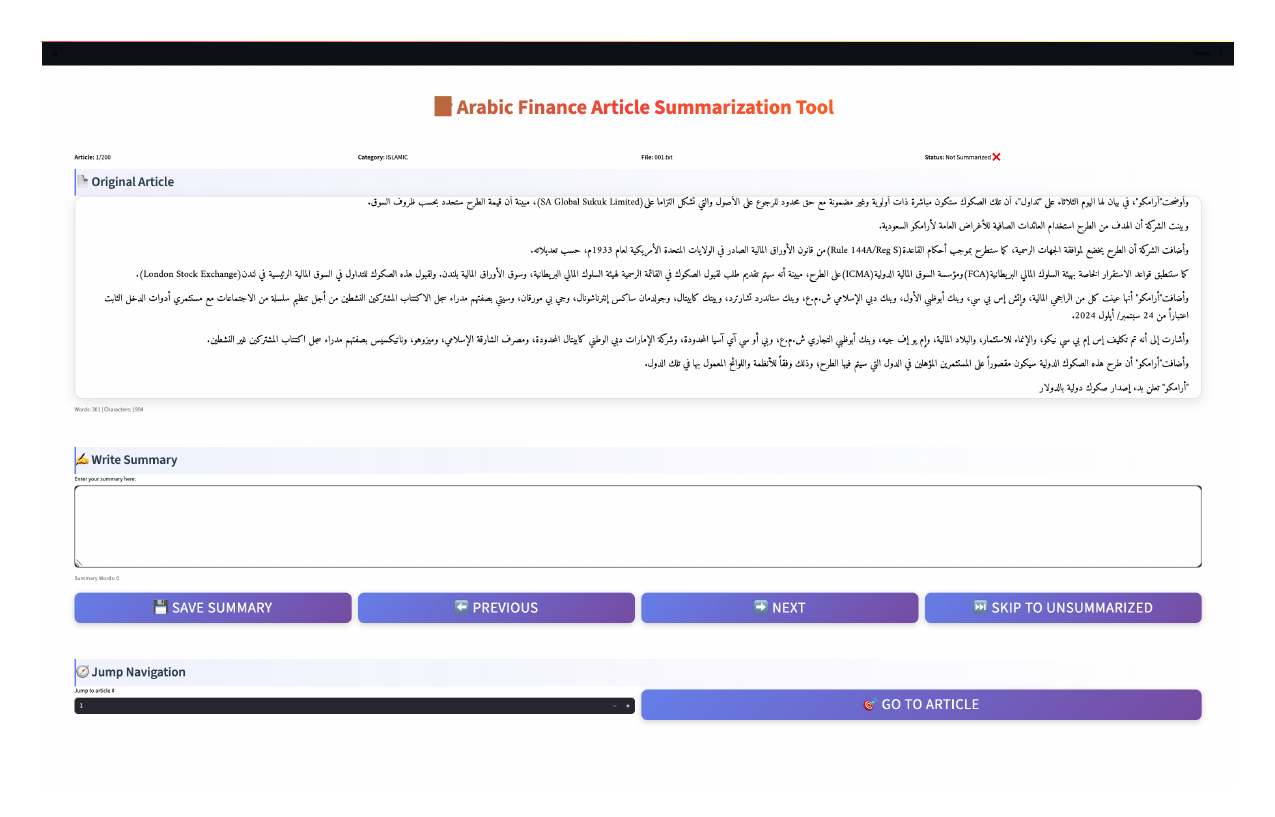}
  \caption{Custom web-based annotation interface for extractive summarization. Annotators view Arabic financial reports, select key sentences containing figures, decisions, and disclosures, and mark them for gold-standard summaries.}
  \label{fig:summarization_annotation}
\end{figure*}

We annotate Arabic financial reports for extractive summarization using a structured human-in-the-loop workflow supported by a custom web-based interface (Figure~\ref{fig:summarization_annotation}) and guided by explicit annotation criteria (Figure~\ref{fig:summarization_guidelines}). This setup standardizes the annotation process, reduces subjectivity, and ensures consistent selection of salient content across annotators. Arabic financial reports also exhibit recurring linguistic and formatting challenges, including specialized terminology, frequent code-switching between Arabic and English, and mixed numeral systems (Table~\ref{tab:arabic_text_difficulties}), all of which complicate sentence selection. Annotators are therefore instructed to carefully identify and select sentences that preserve key financial facts, numerical values, and regulatory references, while strictly avoiding paraphrasing, abstraction, or omission of critical details to maintain fidelity to the source text.

\begin{table}[t]
\centering
\small
\begin{tabular}{lr}
\toprule

\textbf{Category} & \textbf{Total Count} \\
\midrule
Zakat (\foreignlanguage{arabic}{زكاة})         & 4,888 \\
Riba (\foreignlanguage{arabic}{ربا})           & 2,454 \\
Murabaha (\foreignlanguage{arabic}{مرابحة})    & 1,389 \\
Gharar (\foreignlanguage{arabic}{غرر})         & 860 \\
Waqf (\foreignlanguage{arabic}{وقف})           & 730 \\
Ijara (\foreignlanguage{arabic}{إجارة})        & 571 \\
Maysir (\foreignlanguage{arabic}{ميسر})        & 372 \\
Musharaka (\foreignlanguage{arabic}{مشاركة})   & 242 \\
Mudharaba (\foreignlanguage{arabic}{مضاربة})   & 228 \\
Takaful (\foreignlanguage{arabic}{تكافل})      & 187 \\
Sukuk (\foreignlanguage{arabic}{صكوك})         & 32 \\
\midrule
\textbf{Total records} & \textbf{11,953} \\
\bottomrule
\end{tabular}
\caption{Distribution of questions across Islamic finance categories in the final dataset.}
\label{tab:category_counts}
\end{table}

\begin{table}[t]
\centering
\small
\begin{tabularx}{\columnwidth}{>{\raggedright\arraybackslash}p{0.3\columnwidth} X}
\toprule
\rowcolor{softgray}

\textbf{Issue} & \textbf{Example from the report} \\
\midrule
Islamic Finance Terminology &
\foreignlanguage{arabic}{``تعزيز التمويل المستدام وتطوير الصكوك والسندات''} \\
Code-switching &
\foreignlanguage{arabic}{``تستهدف تعزيز التمويل المستدام وتطوير الصكوك والسندات، وزيادة شفافية القطاع \ldots''} \textbf{Fitch} \\
Mixed Numeral Systems &
\foreignlanguage{arabic}{``انخفض حجم الدين بنحو ٢٧  ريال قطري (٧٫٤ مليار دولار) في عام ٢٠٢٣''} — combines Arabic currency and Western digits \\
\bottomrule
\end{tabularx}
\caption{Key text difficulties in Arabic financial reports with real examples}
\label{tab:arabic_text_difficulties}
\end{table}

\begin{figure}[!h]
\centering
\begin{tcolorbox}[title={Extractive Summarization Annotation Guidelines}, width=\linewidth]
\small

\textbf{Task overview.}  
Annotators create extractive summaries by selecting the most important sentences \emph{verbatim} from each Arabic financial document. The goal is to produce a concise summary that preserves critical financial information and reflects the document’s main message.

\vspace{2pt}
\textbf{Document-level assessment.}
\begin{itemize}[noitemsep, topsep=2pt, leftmargin=*]
  \item Read the entire document before selecting any sentences.
  \item Identify the document type (e.g., earnings report, regulatory announcement, market analysis, company news).
  \item Segment the text into sentences using Arabic punctuation marks (\foreignlanguage{arabic}{،} \foreignlanguage{arabic}{؛} \foreignlanguage{arabic}{.}).
  \item Target a summary length of approximately 30--40\% of the original document.
\end{itemize}

\vspace{2pt}
\textbf{Critical content to prioritize.}
Annotators must include sentences containing:
\begin{itemize}[noitemsep, topsep=2pt, leftmargin=*]
  \item \textbf{Financial figures}: \foreignlanguage{arabic}{الأرباح} / \foreignlanguage{arabic}{الخسائر} (profits/losses), \foreignlanguage{arabic}{الإيرادات} (revenues), \foreignlanguage{arabic}{النسب المئوية} (percentages).
  \item \textbf{Performance indicators}: \foreignlanguage{arabic}{نمو} / \foreignlanguage{arabic}{انخفاض} (growth/decline), \foreignlanguage{arabic}{ارتفاع} / \foreignlanguage{arabic}{هبوط} (increase/decrease).
  \item \textbf{Strategic decisions}: \foreignlanguage{arabic}{الاستحواذ} (acquisition), \foreignlanguage{arabic}{الاندماج} (merger), \foreignlanguage{arabic}{التوسع} (expansion).
  \item \textbf{Regulatory or official actions}: \foreignlanguage{arabic}{قرارات الهيئة} (authority decisions), \foreignlanguage{arabic}{الموافقات} (approvals), \foreignlanguage{arabic}{التراخيص} (licenses).
\end{itemize}

\vspace{2pt}
\textbf{Sentence scoring.}
Each sentence is scored on a 1--5 scale:
\begin{itemize}[noitemsep, topsep=2pt, leftmargin=*]
  \item \textbf{5}: Critical financial data or main announcement.
  \item \textbf{4}: Important context or cause--effect explanation.
  \item \textbf{3}: Supporting detail required for clarity.
  \item \textbf{2}: General market or background information.
  \item \textbf{1}: Redundant or generic statements.
\end{itemize}

\vspace{2pt}
\textbf{Selection procedure.}
\begin{itemize}[noitemsep, topsep=2pt, leftmargin=*]
  \item Select all sentences scored \textbf{5}.
  \item Add sentences scored \textbf{4} until the target length is reached.
  \item Include a sentence scored \textbf{3} only if necessary for coherence.
\end{itemize}

\vspace{2pt}
\textbf{Final validation.}
Before submission, annotators verify that the summary:
\begin{itemize}[noitemsep, topsep=2pt, leftmargin=*]
  \item Includes all key financial figures and the main announcement.
  \item Is coherent and understandable on its own.
  \item Falls within the 30--40\% length target.
  \item Avoids repetition and generic background content.
\end{itemize}

\vspace{2pt}
\textbf{Common errors to avoid.}
\begin{itemize}[noitemsep, topsep=2pt, leftmargin=*]
  \item Selecting sentences based solely on position in the document.
  \item Omitting numerical or regulatory information.
  \item Including repetitive or stylistic filler content.
  \item Exceeding the target summary length without justification.
\end{itemize}

\end{tcolorbox}
\caption{Guidelines for extractive summarization annotation of Arabic financial reports.}
\label{fig:summarization_guidelines}
\end{figure}

\section{Event--Cause Reasoning Annotation Guidelines}
\label{app:event_cause_guidelines}


We construct event--cause reasoning instances for Arabic financial reports using a structured annotation framework with explicit quality control procedures (Figure~\ref{fig:event_cause_guidelines}) to ensure consistency, accuracy, and reliable causal interpretation.

\begin{figure*}[!h]
\centering
\begin{tcolorbox}[title={Event--Cause Reasoning QA Annotation and Quality Control}, width=\linewidth]
\small

\textbf{Task objective.}  
Annotators construct one event--cause reasoning instance per Arabic financial report. Each instance consists of (i) an analytical question that requires causal or interpretive reasoning and (ii) a concise expert-written answer grounded exclusively in the information provided in the report. The task evaluates whether models can explain \emph{why} financial or regulatory events occurred and \emph{what their implications are}, rather than recalling isolated facts.

\vspace{2pt}
\textbf{Question construction.}
The question must:
\begin{itemize}[noitemsep, topsep=2pt, leftmargin=*]
  \item Be analytical in nature (e.g., ``why did this occur?'' or ``what does this indicate?'').
  \item Connect multiple data points from the report (e.g., financial figures, growth rates, market reactions).
  \item Avoid purely descriptive prompts (e.g., ``what was the profit?'').
  \item Be answerable using only information stated or implied in the report.
\end{itemize}

\vspace{4pt}
\textbf{Answer construction.}
The answer must:
\begin{itemize}[noitemsep, topsep=2pt, leftmargin=*]
  \item Be written in Arabic and remain concise.
  \item Rely exclusively on the content of the report, without external knowledge or speculation.
  \item Explicitly reference numerical figures and percentages when available.
  \item Provide economic or financial interpretation (e.g., performance drivers, risk implications, or market significance).
  \item Preserve technical and domain-specific terminology.
\end{itemize}

\vspace{4pt}
\textbf{Focus areas.}
Annotators prioritize questions involving:
\begin{itemize}[noitemsep, topsep=2pt, leftmargin=*]
  \item Market trend analysis and its implications.
  \item Performance comparison between companies or sectors.
  \item Economic significance of observed data patterns.
  \item Risk assessment based on reported financial indicators.
\end{itemize}

\vspace{2pt}
\textbf{Quality control procedure.}
We enforce quality control through a multi-stage human validation workflow:

\begin{enumerate}[noitemsep, topsep=2pt, leftmargin=*]
  \item \textbf{Pilot annotation.}  
  Two native Arabic financial experts independently annotate a pilot subset of 20 reports, each producing an event--cause question and an analytical answer.

  \item \textbf{Agreement assessment.}  
  We evaluate agreement at two complementary levels:
  \begin{itemize}[noitemsep, topsep=2pt, leftmargin=*]
    \item \emph{Event--cause identification}: measured using Cohen’s $\kappa$, assessing consistency in identifying salient events and their causes.
    \item \emph{Answer consistency}: measured using ROUGE overlap between independently written answers, used as a consistency check rather than a correctness metric.
  \end{itemize}

  \item \textbf{Calibration.}  
  Annotators review disagreements from the pilot phase, discuss ambiguous cases (e.g., implicit causality, multi-factor events, overlapping economic drivers), and refine shared annotation criteria. This calibration aligns interpretation standards and reduces annotation drift.

  \item \textbf{Full annotation.}  
  After calibration, one expert annotates the remaining reports under the agreed guidelines.

  \item \textbf{Audit and correction.}  
  A senior annotator audits a random sample of completed annotations to verify that each instance:
  \begin{itemize}[noitemsep, topsep=2pt, leftmargin=*]
    \item Identifies a plausible event and its cause(s) supported by the report.
    \item Includes relevant numerical evidence when available.
    \item Provides an analytical explanation rather than a descriptive summary.
  \end{itemize}
  Annotations that fail these checks are revised or discarded.
\end{enumerate}

\vspace{2pt}
\textbf{Final dataset format.}  
Each finalized instance consists of a financial report, one analytical event--cause question, and one expert-written answer. This format supports evaluation using both exact-match and partial-match metrics and enables controlled benchmarking of causal reasoning in Arabic financial text.

\end{tcolorbox}
\caption{Guidelines and quality control workflow for event--cause reasoning annotation in Arabic financial reports.}
\label{fig:event_cause_guidelines}
\end{figure*}


\section{MCQ Answer Normalization and Scoring}
\label{app:mcq_normalization}

To ensure fair and reproducible evaluation of multiple-choice questions, we normalize model outputs before computing accuracy. Large language models frequently generate free-form responses (e.g., explanations, mixed scripts, or multiple answer mentions) rather than a single option label.

\begin{figure}[!h]
\centering
\begin{tcolorbox}[title={Prompt for Extracting Arabic Accounting Exam MCQs (Exercise-Based Format)}, width=\linewidth]
\small

\textbf{Task.}
You are an expert system for extracting Arabic accounting exam questions. The input is a scanned exam page containing exercises that begin with the keyword \foreignlanguage{arabic}{تمرين} (Exercise), followed by a number.

\vspace{2pt}
\textbf{Extraction instructions.}
\begin{itemize}[noitemsep, topsep=2pt, leftmargin=*]
  \item Identify all exercises that begin with \foreignlanguage{arabic}{تمرين} followed by a numeral (e.g., \foreignlanguage{arabic}{تمرين ١}, \foreignlanguage{arabic}{تمرين ٢}).
  \item For each exercise, extract the full contextual text that follows the exercise header.
  \item Within each exercise, identify all multiple-choice questions (numbered 1, 2, 3, etc.).
  \item For each MCQ, combine the exercise context with the specific question text to form a complete question.
  \item Extract all answer choices labeled as \foreignlanguage{arabic}{أ}, \foreignlanguage{arabic}{ب}, \foreignlanguage{arabic}{ج}, and \foreignlanguage{arabic}{د}.
  \item Identify the correct answer by detecting underlined text in the choices; underlining indicates the correct option.
\end{itemize}

\vspace{2pt}
\textbf{Critical rules.}
\begin{itemize}[noitemsep, topsep=2pt, leftmargin=*]
  \item Preserve the original Arabic text exactly; do not paraphrase or normalize.
  \item Extract \emph{all} MCQs appearing under each exercise.
  \item If no underlined choice is visible, set the correct answer to \texttt{null}.
\end{itemize}

\vspace{2pt}
\textbf{Output format.}
Return a JSON object with the following structure:
\begin{verbatim}
{
  "exercises": [
    {
      "exercise_number": "...",
      "exercise_context": "...",
      "questions": [
        {
          "question_number": "...",
          "full_question_text": "...",
          "choices": {
            "أ": "...",
            "ب": "...",
            "ج": "...",
            "د": "..."
          },
          "correct_answer": "...",
          "is_underlined": true
        }
      ]
    }
  ],
  "page_info": {
    "total_exercises": ...,
    "total_questions": ...,
    "language": "Arabic",
    "subject": "Accounting"
  }
}
\end{verbatim}

\end{tcolorbox}
\caption{Prompt for extracting MCQs from Arabic accounting exams with exercise-based layouts.}
\label{fig:accounting_prompt}
\end{figure}

\begin{figure}[!h]
\centering
\begin{tcolorbox}[title={Prompt for Extracting Arabic Business Exam MCQs (Tabular Format)}, width=\linewidth]
\small

\textbf{Task.}
You are an expert system for extracting Arabic business and accounting exam questions from scanned images containing tabular layouts.

\vspace{2pt}
\textbf{Document characteristics.}
\begin{itemize}[noitemsep, topsep=2pt, leftmargin=*]
  \item Each row corresponds to one question.
  \item Questions are numbered using Arabic numerals (e.g., \foreignlanguage{arabic}{١٢٤}, \foreignlanguage{arabic}{١٢٥}).
  \item Answer choices typically include \foreignlanguage{arabic}{صح} (True) and \foreignlanguage{arabic}{خطأ} (False), with optional additional choices.
  \item The correct answer is highlighted with a yellow background.
\end{itemize}

\vspace{2pt}
\textbf{Extraction instructions.}
\begin{itemize}[noitemsep, topsep=2pt, leftmargin=*]
  \item Identify all question rows in the table.
  \item Extract the question number and full Arabic question text for each row.
  \item Extract all visible answer choices.
  \item Identify the correct answer by detecting yellow highlighting.
  \item Label questions as \texttt{true\_false} or \texttt{multiple\_choice} accordingly.
\end{itemize}

\vspace{2pt}
\textbf{Critical rules.}
\begin{itemize}
  \item Preserve Arabic text exactly as written, including diacritics.
  \item If yellow highlighting is ambiguous or not visible, set \texttt{has\_yellow\_highlight} to \texttt{false}.
  \item Extract \emph{all} visible questions on the page.
\end{itemize}

\vspace{2pt}
\textbf{Output format.}
Return a JSON object with the following structure:
\begin{verbatim}
{
  "questions": [
    {
      "question_number": "...",
      "question_text": "...",
      "question_type": "...",
      "choices": {
        "a": "...",
        "b": "...",
        "c": "..."
      },
      "correct_answer": "...",
      "correct_choice_text": "...",
      "has_yellow_highlight": true,
      "subject_area": "business"
    }
  ],
  "page_info": {
    "total_questions": ...,
    "format": "table_with_yellow_highlighting",
    "language": "Arabic",
    "question_type": "mixed"
  }
}
\end{verbatim}

\end{tcolorbox}
\caption{Prompt for extracting MCQs from Arabic business and accounting exams with tabular layouts.}
\label{fig:business_prompt}
\end{figure}

\paragraph{Normalization procedure.}
For each model output, we apply the following steps:
\begin{itemize}
  \item Normalize Unicode and Arabic script by removing diacritics, collapsing repeated whitespace and punctuation, and mapping Eastern Arabic digits (e.g., \foreignlanguage{arabic}{١٢٣٤}) to Western digits (\texttt{1234}).
  \item Extract the first explicit answer mention using a cascade of regular expressions that handle:
  \begin{itemize}
    \item Latin option labels (e.g., \texttt{A}, \texttt{B}, ``Option C''),
    \item Arabic option letters (e.g., \foreignlanguage{arabic}{أ}, \foreignlanguage{arabic}{ب}, \foreignlanguage{arabic}{ج}),
    \item Spelled-out Arabic forms (e.g., \foreignlanguage{arabic}{باء}),
    \item Numeric indices (e.g., \texttt{1--4}).
  \end{itemize}
\end{itemize}

\paragraph{Scoring.}
We compute accuracy as an exact match between the normalized prediction $\hat{y}$ and the gold label $y$. For example, the output ``\foreignlanguage{arabic}{الإجابة هي 2 بسبب صياغة الحكم}'' is normalized to \texttt{B}, while ``\foreignlanguage{arabic}{الخيار (ج) هو الصحيح}'' is normalized to \texttt{C}. Outputs that do not contain a valid option after normalization are marked incorrect. This procedure ensures that evaluation is robust to superficial variation in formatting, language mixing, and numeral systems, and that all models are assessed under a consistent and deterministic scoring protocol.

\section{Instruction Templates for SAHM Tasks}
\label{app:instruction_templates}

To enable a unified instruction-tuning and evaluation setup across heterogeneous tasks, we convert each SAHM task into a standardized instruction format. Table~\ref{tab:sahm_instruction_templates} lists the canonical task instructions used in our benchmark, shown in their original Arabic formulation alongside an English translation for clarity. The Arabic prompts constitute the actual inputs used during model evaluation, while the English versions are provided solely to document task intent and facilitate reproducibility.

\begin{table*}[!h]
\centering
\small
\setlength{\tabcolsep}{2pt}
\renewcommand{\arraystretch}{1.3}

\begin{tabular}{@{}p{0.22\linewidth} p{0.39\linewidth} p{0.39\linewidth}@{}}
\toprule
\textbf{Dataset} & \textbf{Original Arabic Prompt} & \textbf{English Translated Prompt} \\
\midrule

\texttt{Islamic Sharia Standards QA}
&
\foreignlanguage{arabic}{بناءً على معايير وأحكام التمويل الإسلامي والمعاملات المالية الشرعية، أجب على السؤال التالي بدقة.}
\foreignlanguage{arabic}{ \textbf{Text:}  السؤال: \{Question\}.}
\foreignlanguage{arabic}{ \textbf{الإجابة وفقاً للضوابط الشرعية:} \{Output\}}
&
Based on Islamic finance standards and Shari’ah-compliant rulings, answer the following question accurately.
\textbf{Text:}  Question: \{Question\}.
\textbf{Answer (Shari’ah-compliant):} \{Output\}
\\
\midrule

\texttt{Islamic Fatwa QA}
&
\foreignlanguage{arabic}{بناءً على أحكام الشريعة الإسلامية والفقه الإسلامي، أجب على السؤال التالي بطريقة مفصلة ومدعمة بالأدلة عند الإمكان.}
\foreignlanguage{arabic}{ \textbf{Text:} السؤال: \{Question\}.}
\foreignlanguage{arabic}{ \textbf{Answer:} \{Output\}}
&
Based on Islamic jurisprudence (fiqh) and Shari’ah rulings, answer the following question in a detailed manner, supported by evidence when possible.
\textbf{Text:} Question: \{Question\}.
\textbf{Answer:} \{Output\}
\\
\midrule

\texttt{Islamic Financial Fatwa MCQ}
&
\foreignlanguage{arabic}{اقرأ السؤال التالي بعناية واختر الإجابة الصحيحة وفقاً لأحكام الشريعة.}
\foreignlanguage{arabic}{ \textbf{Text:} السؤال: \{Question\}. الخيارات: \{Choices\}.}
\foreignlanguage{arabic}{ \textbf{Answer:} أخرج حرف الخيار الصحيح فقط.}
&
Read the following question carefully and choose the correct answer according to Shari’ah rulings.
\textbf{Text:} Question: \{Question\}. Choices: \{Choices\}.
\textbf{Answer:} Output only the correct option letter.
\\
\midrule

\texttt{Accounting Exams MCQ}
&
\foreignlanguage{arabic}{اقرأ السؤال التالي بعناية واختر الإجابة الصحيحة.}
\foreignlanguage{arabic}{ \textbf{Text:} السؤال: \{Question\}. الخيارات: \{Choices\}.}
\foreignlanguage{arabic}{ \textbf{Answer:} أخرج حرف الخيار الصحيح فقط.}
&
Read the following question carefully and choose the correct answer.
\textbf{Text:} Question: \{Question\}. Choices: \{Choices\}.
\textbf{Answer:} Output only the correct option letter.
\\
\midrule

\texttt{Business Exams MCQ }
&
\foreignlanguage{arabic}{اقرأ السؤال التالي بعناية واختر الإجابة الصحيحة.}
\foreignlanguage{arabic}{ \textbf{Text:} السؤال: \{Question\}. الخيارات: \{Choices\}.}
\foreignlanguage{arabic}{ \textbf{Answer:} أخرج حرف الخيار الصحيح فقط.}
&
Read the following business/management question carefully and choose the correct answer.
\textbf{Text:} Question: \{Question\}. Choices: \{Choices\}.
\textbf{Answer:} Output only the correct option letter.
\\
\midrule

\texttt{Financial Report Sentiment Analysis MCQ}
&
\foreignlanguage{arabic}{اقرأ بعناية التقرير المالي التالي واختر التصنيف الصحيح من منظور المستثمر.}
\foreignlanguage{arabic}{ \textbf{Text:} التقرير: \{Input\}.}
\foreignlanguage{arabic}{ \textbf{Answer:} (إيجابي / سلبي / محايد).}
&
Read the following financial report carefully and choose the correct label from an investor's perspective.
\textbf{Text:} Report: \{Input\}.
\textbf{Answer:} (Positive / Negative / Neutral).
\\
\midrule

\texttt{Report Extractive Summarization}
&
\foreignlanguage{arabic}{قم بتلخيص التقرير المالي التالي باستخدام التلخيص الاستخراجي (Extractive Summarization).}
\foreignlanguage{arabic}{اختر الجمل الأكثر أهمية مباشرة من النص الأصلي دون تعديل أو إعادة صياغة، ورتّبها بنفس تسلسلها.}
\foreignlanguage{arabic}{اجعل الملخص حوالي 30--40\% من حجم النص، وركّز على الأرقام والقرارات والنتائج والتواريخ.}
\foreignlanguage{arabic}{ \textbf{Text:} التقرير: \{Input\}.}
\foreignlanguage{arabic}{ \textbf{Answer:} أخرج الملخص فقط دون أي شرح.}
&
Summarize the following financial report using extractive summarization (select sentences verbatim, keep original order, target 30--40\% length, focus on numbers/decisions/outcomes/dates).
\textbf{Text:} Report: \{Input\}.
\textbf{Answer:} Output the extractive summary only (no extra text).
\\
\midrule

\texttt{Event–Cause Reasoning QA }
&
\foreignlanguage{arabic}{بناءً على التقرير المالي التالي، أجب على السؤال التحليلي بشكل مفصل ودقيق مع الالتزام بالمعلومات الواردة في النص فقط.}
\foreignlanguage{arabic}{ \textbf{Text:} التقرير المالي: \{Input\}. السؤال: \{Question\}.}
\foreignlanguage{arabic}{ \textbf{Answer:} \{Output\}}
&
Based on the following financial report, answer the analytical question in a detailed and accurate way, grounded only in the provided text.
\textbf{Text:} Financial report: \{Input\}. Question: \{Question\}.
\textbf{Answer:} \{Output\}
\\

\bottomrule
\end{tabular}
\caption{Instruction templates used for SAHM tasks (Arabic prompts are used in evaluation; English translations document task intent).}
\label{tab:sahm_instruction_templates}
\end{table*}

\section{LLM-as-a-Judge Protocol, Validation, and Reproducibility}
\label{app:judge_rubrics}

\subsection{Judge Protocol and Reproducibility}
\label{app:judge_protocol}

We evaluate the three open-ended tasks (Fatwa QA, Shari’ah Standards QA, and Event--Cause QA) using an LLM-as-a-judge setup with \texttt{Gemini-2.5-Flash}. For each instance, the judge receives: (i) the Arabic prompt (including any report/excerpt and question), (ii) the gold reference answer, and (iii) the model's candidate answer. 

The judge is blind to model identity and observes inputs in fixed, labeled fields (\texttt{prompt}, \texttt{ground\_truth}, \texttt{candidate\_answer}) to avoid positional ambiguity. It returns a structured JSON with: (a) rubric sub-scores summing to $[0,10]$, (b) task-specific critical error flags (e.g., contradiction with the reference, omission of critical constraints, normalization of unlawful elements, or fabrication/alteration of figures), and (c) a brief explanation. Task-specific evaluation rubrics are defined for fatwa QA (Figure~\ref{fig:fatwa_judge}), Islamic finance QA (Figure~\ref{fig:islamic_judge}), and financial analysis tasks (Figure~\ref{fig:finance_judge}).


We enforce a strict JSON schema during parsing to ensure all judge outputs are machine-readable and consistently structured. If a response is invalid JSON or violates the schema, we retry once with the same inputs and an explicit \emph{JSON-only} instruction to correct formatting issues without altering content. Persistent failures are marked invalid and excluded from aggregate scores, and we report the invalid-rate as a transparency measure. We run the judge deterministically (temperature $=0.0$, greedy decoding, max output tokens $=4096$), eliminating randomness. 

As a result, we do not perform repeated judging or score averaging. This ensures reproducibility and consistency across all evaluations, yielding stable outputs under fixed inputs and identical evaluation conditions.

\subsection{Human Alignment Study (Judge Validation)}
\label{app:judge_validation}

To validate the LLM judge against expert evaluation, we conduct a human alignment study on $200$ randomly sampled open-ended outputs spanning Fatwa QA, Shari’ah Standards QA, and Event--Cause QA. Two expert Arabic annotators independently score each model response using the same $[0,10]$ additive rubric provided to the judge (Section~\ref{app:judge_rubrics}). This setup enables direct comparison between human and model-based scoring under identical evaluation criteria.


We compare the judge’s scores (\texttt{Gemini-2.5-Flash}) to the mean human scores, obtaining an MSE of $0.41$ and Pearson $r=0.92$, indicating strong alignment. Inter-annotator agreement is high ($\kappa=0.84$ computed on discretized integer scores), demonstrating consistent agreement across annotators.

These results indicate that the LLM-as-a-judge scores closely track expert human judgments under our rubric, supporting its reliability as an evaluation proxy.

\subsection{Cross-Judge Validation of Open-Ended Evaluations}
\label{app:cross_judge}


To address concerns about potential model-family bias in our LLM-as-judge evaluation, we re-ran all open-ended tasks with two independent judges: Gemini-2.5-Flash (primary) and GPT-4o.

Each model was evaluated 3 times under greedy decoding, and we report mean$\pm$std across runs for both judges, ensuring robustness and consistency of evaluation results. If our primary judge were favoring Gemini-family models, switching to GPT-4o should lower their scores; instead, they \emph{rise} (Gemini-3-Flash: Islamic-Std 9.18$\to$9.76, Fatwa 9.17$\to$9.32), the opposite of circular bias. This consistency supports the robustness of our evaluation setup.\\\\

Model rankings are preserved across judges: top-tier models (Gemini-3-Flash, Claude-Opus-4.5) and bottom-tier models (SILMA-9B, LLaMA-3.1-8B) remain in the same groupings regardless of judge. Tight confidence intervals ($\pm$0.02--0.10) across 3 runs confirm reproducibility under greedy decoding. Full per-model, per-judge scores appear in Table~\ref{tab:cross_judge}. These results indicate stable evaluation outcomes across different judge models.

\begin{table*}[h]
\centering
\small
\begin{tabular}{lcccccc}
\toprule
& \multicolumn{2}{c}{\textbf{Event-Cause QA}} & \multicolumn{2}{c}{\textbf{Islamic-Std QA}} & \multicolumn{2}{c}{\textbf{Fatwa QA}} \\
\cmidrule(lr){2-3} \cmidrule(lr){4-5} \cmidrule(lr){6-7}
\textbf{Model} & \textbf{Gemini Judge} & \textbf{GPT-4o Judge} & \textbf{Gemini Judge} & \textbf{GPT-4o Judge} & \textbf{Gemini Judge} & \textbf{GPT-4o Judge} \\
\midrule
Gemini-3-Flash        & 9.84$\pm$0.02 & 9.97$\pm$0.02 & 9.18$\pm$0.03 & 9.76$\pm$0.01 & 9.17$\pm$0.02 & 9.32$\pm$0.02 \\
Claude-Opus-4.5       & 9.67$\pm$0.03 & 9.32$\pm$0.97 & 8.06$\pm$0.05 & 9.53$\pm$0.02 & 8.79$\pm$0.03 & 9.18$\pm$0.02 \\
Claude-Sonnet-4.5     & 9.32$\pm$0.04 & 9.68$\pm$0.04 & 8.24$\pm$0.05 & 9.28$\pm$0.02 & 7.58$\pm$0.03 & 8.86$\pm$0.02 \\
GPT-4o                & 8.30$\pm$0.06 & 9.50$\pm$0.09 & 6.64$\pm$0.08 & 8.53$\pm$0.03 & 6.50$\pm$0.04 & 8.04$\pm$0.01 \\
Gemma-3-27B           & 8.70$\pm$0.05 & 9.74$\pm$0.02 & 6.15$\pm$0.08 & 8.41$\pm$0.03 & 5.18$\pm$0.05 & 7.35$\pm$0.02 \\
Qwen2.5-72B           & 8.08$\pm$0.10 & 8.45$\pm$0.17 & 5.61$\pm$0.10 & 6.96$\pm$0.07 & 5.37$\pm$0.06 & 6.30$\pm$0.01 \\
Fanar-1-9B            & 7.57$\pm$0.10 & 8.03$\pm$0.22 & 4.94$\pm$0.08 & 6.03$\pm$0.03 & 4.44$\pm$0.06 & 5.15$\pm$0.04 \\
Gemma-3-4B            & 7.39$\pm$0.08 & 9.09$\pm$0.08 & 2.88$\pm$0.08 & 5.32$\pm$0.07 & 2.46$\pm$0.06 & 4.39$\pm$0.02 \\
ALLAM-7B              & 6.87$\pm$0.10 & 7.79$\pm$0.16 & 4.92$\pm$0.08 & 5.95$\pm$0.03 & 4.20$\pm$0.05 & 4.49$\pm$0.03 \\
LLaMA-3.1-70B         & 6.60$\pm$0.15 & 7.15$\pm$0.30 & 3.70$\pm$0.10 & 4.67$\pm$0.05 & 4.74$\pm$0.08 & 2.24$\pm$0.02 \\
Mixtral-8x7B          & 4.53$\pm$0.08 & 5.14$\pm$0.09 & 2.48$\pm$0.08 & 3.43$\pm$0.08 & 1.78$\pm$0.06 & 2.80$\pm$0.04 \\
SILMA-9B              & 1.88$\pm$0.20 & 1.43$\pm$0.37 & 3.33$\pm$0.12 & 2.33$\pm$0.03 & 2.05$\pm$0.08 & 1.61$\pm$0.05 \\
LLaMA-3.1-8B          & 4.90$\pm$0.18 & 2.49$\pm$0.17 & 2.50$\pm$0.12 & 1.85$\pm$0.03 & 1.38$\pm$0.08 & 0.71$\pm$0.02 \\
\midrule
\textsc{Sahm-ALLAM-7B} & 6.50$\pm$0.10 & 7.02$\pm$0.12 & 6.30$\pm$0.02 & 6.59$\pm$0.04 & 4.24$\pm$0.04 & 4.51$\pm$0.03 \\
\bottomrule
\end{tabular}
\caption{Cross-judge validation on open-ended tasks with Gemini-2.5-Flash and GPT-4o (mean$\pm$std over 3 runs). Rankings are preserved, and Gemini scores rise under GPT-4o, indicating no circular bias.}
\label{tab:cross_judge}
\end{table*}

\subsection{Frontier Model Error Analysis}
\label{app:frontier_errors}

\begin{table*}[!h]
\centering
\small
\begin{tabular}{lcccc}
\toprule
\textbf{Model} & \textbf{Accounting (\%)} & \textbf{Business (\%)} & \textbf{Fatw\=a MCQ (\%)} & \textbf{Sentiment (\%)} \\
\midrule
\rowcolor{softgray}
\multicolumn{5}{c}{\textbf{Proprietary}} \\
\midrule
Claude-Opus-4.5       & 78.04$\pm$2.42 & 76.14$\pm$1.14 & 91.57$\pm$0.33 & 61.25$\pm$2.50 \\
Claude-Sonnet-4.5     & 77.25$\pm$1.20 & 77.05$\pm$1.45 & 88.83$\pm$0.38 & 66.25$\pm$1.25 \\
Gemini-3-Flash        & 74.65$\pm$1.95 & 75.41$\pm$0.95 & 90.07$\pm$0.30 & 70.00$\pm$1.25 \\
GPT-5                 & 63.67$\pm$2.27 & 72.31$\pm$1.26 & 91.15$\pm$0.45 & 62.50$\pm$1.25 \\
GPT-4o                & 59.28$\pm$2.07 & 78.32$\pm$0.32 & 87.50$\pm$0.10 & 61.25$\pm$0.00 \\
Gemini-2.5-Flash      & 55.49$\pm$2.13 & 75.05$\pm$0.83 & 86.02$\pm$1.22 & 58.33$\pm$4.39 \\
\midrule
\rowcolor{softgray}
\multicolumn{5}{c}{\textbf{Open-source $\geq$70B}} \\
\midrule
Qwen2.5-72B           & 63.08$\pm$2.70 & 75.23$\pm$0.32 & 83.63$\pm$0.33 & 64.00$\pm$1.25 \\
LLaMA-3.1-70B         & 49.11$\pm$2.79 & 75.58$\pm$1.14 & 82.90$\pm$0.15 & 51.25$\pm$3.31 \\
\midrule
\rowcolor{softgray}
\multicolumn{5}{c}{\textbf{Open-source $<$70B}} \\
\midrule
Gemma-3-27B           & 53.29$\pm$2.16 & 74.13$\pm$0.32 & 80.67$\pm$0.18 & 64.17$\pm$0.72 \\
Gemma-2-9B            & 46.71$\pm$2.74 & 65.31$\pm$3.83 & 70.43$\pm$0.61 & 54.17$\pm$1.44 \\
Qwen2.5-14B           & 48.49$\pm$3.93 & 64.66$\pm$0.83 & 75.18$\pm$0.85 & 60.83$\pm$3.82 \\
Qwen2.5-7B            & 46.11$\pm$2.85 & 63.02$\pm$1.14 & 69.70$\pm$0.28 & 54.17$\pm$1.91 \\
Gemma-3-4B            & 38.12$\pm$2.27 & 67.58$\pm$0.32 & 61.30$\pm$0.18 & 62.08$\pm$1.44 \\
Mixtral-8x7B          & 31.74$\pm$1.04 & 59.38$\pm$0.63 & 62.32$\pm$0.34 & 58.33$\pm$0.72 \\
LLaMA-3.1-8B          & 38.93$\pm$3.28 & 58.64$\pm$4.45 & 60.35$\pm$3.62 & 52.08$\pm$5.77 \\
\midrule
\rowcolor{softgray}
\multicolumn{5}{c}{\textbf{Arabic Models}} \\
\midrule
Fanar-1-9B            & 43.51$\pm$2.42 & 70.13$\pm$1.67 & 74.60$\pm$0.35 & 60.42$\pm$2.60 \\
SILMA-9B              & 49.32$\pm$21.73 & 60.11$\pm$6.61 & 53.85$\pm$5.57 & 25.75$\pm$3.75 \\
ALLAM-7B              & 42.24$\pm$3.55 & 64.75$\pm$3.83 & 72.25$\pm$2.83 & 56.50$\pm$2.00 \\
\bottomrule
\end{tabular}
\caption{MCQ evaluation across 3 runs using each model's recommended temperature. Values shown as mean$\pm$std. Rankings remain consistent across runs, confirming robustness of our main findings to decoding configuration.}
\label{tab:mcq_variance}
\end{table*}

To diagnose why frontier models fail on specific \name{} tasks, we conducted a systematic error analysis of GPT-5 and Gemini-3-Flash across Accounting, Business, and Summarization. Two native Arabic annotators with financial backgrounds jointly reviewed each error, analyzed the reasoning against the gold reference, assigned a root cause, and agreed on a category (full agreement after adjudication). We use a shared taxonomy across both models: \textbf{Misunderstanding Concept} (correct setup, wrong principle applied), \textbf{Concept Confusion} (conflates two related but distinct concepts), \textbf{Hallucination} (generates facts not in the question), \textbf{Question Misread} (answers a different question), \textbf{Calculation Mistake} (arithmetic error), and \textbf{Domain Knowledge Gap} (lacks the terminology entirely).

\begin{figure*}[!h]
\centering
\begin{tcolorbox}[title={LLM-as-a-Judge Rubric for Fatwa QA (Arabic)}, width=\linewidth]
\small

\textbf{Role.} You are an expert evaluator in Islamic fatwa (iftāʾ).

\vspace{2pt}
\textbf{Inputs (provided each time).}
\begin{itemize}[noitemsep, topsep=2pt, leftmargin=*]
  \item \texttt{category} (optional context) -- may be empty (e.g., riba, zakat, takaful)
  \item \texttt{prompt} -- the full Arabic prompt shown to the model (instructions + question)
  \item \texttt{ground\_truth} -- the reference fatwa answer (Arabic)
  \item \texttt{candidate\_answer} -- the model answer to evaluate (Arabic)
\end{itemize}

\vspace{2pt}
\textbf{Task.}
Judge how well \texttt{candidate\_answer} matches \texttt{ground\_truth} in \emph{ruling (ḥukm)}, \emph{justification}, and \emph{operative constraints/qualifications}. Prioritize doctrinal correctness and required conditions/exceptions. Do not penalize stylistic paraphrase if the core ruling and constraints are preserved. Be concise and deterministic.

\vspace{2pt}
\textbf{Scoring (sum to exactly 10).}
\begin{enumerate}[noitemsep, topsep=2pt, leftmargin=*]
  \item \textbf{Coverage of core ruling (0--4).}
  The candidate must clearly state the same central hukm (e.g., permissibility/prohibition, validity/invalidity) \emph{and} include the key justification present in the ground truth. One-word/minimal answers without essential justification should receive a much lower score (e.g., 0--1).
  \item \textbf{Conditions, exceptions, constraints (0--2).}
  Does it retain critical restrictions, qualifiers, or carve-outs that materially affect the ruling?
  \item \textbf{Doctrinal/factual accuracy (0--2).}
  No misstatements that would change the fatwa; no implicit legalization of prohibited elements (e.g., ribā); no misleading generalizations or invented requirements.
  \item \textbf{Clarity \& Arabic language quality (0--1).}
  Clear Arabic, understandable structure, minimal ambiguity appropriate for a fatwa answer.
  \item \textbf{Directness \& fatwa format (0--1).}
  Directly answers the question; avoids long digressions; phrasing suitable for a fatwa.
\end{enumerate}

\vspace{2pt}
\textbf{Critical checks (true/false).}
\begin{itemize}[noitemsep, topsep=2pt, leftmargin=*]
  \item \texttt{contradicts\_ground\_truth}: Does the candidate contradict the central ruling?
  \item \texttt{omits\_critical\_conditions}: Does it omit key conditions/exceptions that change the ruling?
  \item \texttt{introduces\_unlawful\_elements}: Does it introduce/normalize prohibited elements (e.g., ribā)?
  \item \texttt{hallucinated\_citations}: Misleading/fabricated sources claimed that distort the ruling?
  \item \texttt{non\_answer\_or\_evasive}: Does it avoid giving a clear ruling?
  \item \texttt{off\_topic\_or\_unsafe}: Off-topic or otherwise inappropriate?
\end{itemize}

\vspace{2pt}
\textbf{Output format (strict).}
Output \emph{only} valid JSON (no prose, no code fences), following this schema:
\begin{verbatim}
{
  "scores": {
    "coverage_core_ruling": <float 0-4>,
    "conditions_exceptions": <float 0-2>,
    "factual_doctrinal_accuracy": <float 0-2>,
    "clarity_language": <float 0-1>,
    "directness_format": <float 0-1>
  },
  "overall": <float 0-10>,
  "critical_checks": {
    "contradicts_ground_truth": <true/false>,
    "omits_critical_conditions": <true/false>,
    "introduces_unlawful_elements": <true/false>,
    "hallucinated_citations": <true/false>,
    "non_answer_or_evasive": <true/false>,
    "off_topic_or_unsafe": <true/false>
  },
  "note": "<short NOTE in {NOTE_LANG}>"
}
\end{verbatim}

\end{tcolorbox}
\caption{Evaluation rubric used for LLM-based judgment of fatwa QA responses.}
\label{fig:fatwa_judge}
\end{figure*}

\subsubsection{GPT-5 on Accounting}
\label{app:gpt5_accounting}

Seventy percent of GPT-5 accounting errors stem from domain reasoning failures: Misunderstanding Concept (39\%) and Concept Confusion (31\%), while Calculation Mistakes account for only 20\%. This confirms our Section~\ref{sec:error_analysis} finding that models rarely fail at arithmetic but often fail at choosing the correct computation. Concretely, GPT-5 (i) reaches correct intermediate results but selects the wrong accounting standard, (ii) confuses closely related concepts (e.g., treating a direct relationship as inverse), and (iii) in 15\% of errors, reaches the correct answer but fabricates a rule to justify switching to a wrong option.


These patterns indicate the model’s weaknesses lie in conceptual grounding rather than computational ability. Errors arise when mapping problem statements to the correct accounting principle or standard, not during numerical execution. This suggests improving domain-specific reasoning and conceptual alignment is more critical than enhancing raw calculation capabilities for such tasks.

\begin{figure}[!h]
\centering
\begin{tcolorbox}[title={LLM-as-a-Judge Rubric for Islamic Finance QA (Arabic)}, width=\linewidth]
\small

\textbf{Role.} You are an expert evaluator in Islamic finance (\emph{Fiqh al-mu\textquotesingle \=amal\=at}).

\vspace{2pt}
\textbf{Inputs (provided each time).}
\begin{itemize}[noitemsep, topsep=2pt, leftmargin=*]
  \item \texttt{topic} (optional context) -- may be empty
  \item \texttt{question} -- in Arabic
  \item \texttt{ground\_truth} -- the reference correct answer (Arabic)
  \item \texttt{candidate\_answer} -- the model answer to evaluate (Arabic)
\end{itemize}

\vspace{2pt}
\textbf{Task.}
Judge how well \texttt{candidate\_answer} matches \texttt{ground\_truth} in \emph{meaning}, \emph{ruling}, \emph{justification}, and \emph{constraints}. Prioritize doctrinal correctness and completeness of key conditions/exceptions. Do not penalize stylistic paraphrase if the core ruling and constraints are preserved. Be concise and deterministic.

\vspace{2pt}
\textbf{Scoring (sum to exactly 10).}
\begin{enumerate}[noitemsep, topsep=2pt, leftmargin=*]
  \item \textbf{Coverage of core ruling (0--4).}
  \item \textbf{Conditions, exceptions, constraints (0--2).}
  \item \textbf{Doctrinal/factual accuracy (0--2).}
  \item \textbf{Clarity \& Arabic language quality (0--1).}
  \item \textbf{Directness \& on-topic (0--1).}
\end{enumerate}

\vspace{2pt}
\textbf{Critical checks (true/false).}
\begin{itemize}[noitemsep, topsep=2pt, leftmargin=*]
  \item \texttt{contradicts\_ground\_truth}
  \item \texttt{omits\_critical\_conditions}
  \item \texttt{introduces\_unlawful\_elements}
  \item \texttt{hallucinated\_citations}
  \item \texttt{non\_answer\_or\_evasive}
  \item \texttt{off\_topic\_or\_unsafe}
\end{itemize}

\vspace{2pt}
\textbf{Output format (strict).}
Output \emph{only} valid JSON (no prose, no code fences), following this schema:
\begin{verbatim}
{ "scores": {"coverage_core_ruling": <float 0-4>,
    "conditions_exceptions": <float 0-2>,
    "factual_doctrinal_accuracy": <float 0-2>,
    "clarity_language": <float 0-1>,
    "directness_format": <float 0-1>},
  "overall": <float 0-10>,
  "critical_checks": {
    "contradicts_ground_truth": <true/false>,
    "omits_critical_conditions": <true/false>,
    "introduces_unlawful_elements": <true/false>,
    "hallucinated_citations": <true/false>,
    "non_answer_or_evasive": <true/false>,
    "off_topic_or_unsafe": <true/false>
    },
  "note": "<short NOTE in {NOTE_LANG}>" }
\end{verbatim}

\end{tcolorbox}
\caption{Evaluation rubric used for LLM-based judgment of Islamic finance QA responses.}
\label{fig:islamic_judge}
\end{figure}

\subsubsection{GPT-5 on Extractive Summarization}
\label{app:gpt5_summarization}


GPT-5 understands report content but fails at task execution, selecting background sentences over key financial figures (42.5\% of errors), copying entire reports instead of meeting the 30--40\% compression target (16.3\%), and introducing content from unrelated reports (11.3\%). This explains its poor extractive summarization performance (ROUGE-L: 33.37) despite strong open-ended reasoning: the task rewards verbatim selection discipline, not generative fluency.

\subsubsection{Gemini-3-Flash on Accounting}
\label{app:gemini_accounting}


Concept Confusion (27.8\%) and Misunderstanding Concept (19.4\%) account for 47\% of errors, particularly in auditing standards and foreign currency hedging. Unlike GPT-5, Gemini-3-Flash also exhibits Hallucination (8.3\%) and Question Misread (8.3\%), while Calculation Mistakes remain rare (5.6\%). These patterns indicate broader instability beyond core conceptual errors, reflecting less consistent reasoning behavior overall.

\subsubsection{Gemini-3-Flash on Business}
\label{app:gemini_business}

Reasoning Error (39.5\%) and Concept Confusion (37.2\%) dominate at 77\%, concentrated in Strategic Management, Marketing, and Entrepreneurship, which require culturally grounded Arabic business knowledge. Domain Knowledge Gap (9.3\%) reflects cases where models lack specialized Arabic business terminology. These patterns highlight the importance of domain-specific knowledge beyond general language understanding.

\subsubsection{Cross-Model Convergence}
\label{app:cross_model_convergence}


The most striking finding is that GPT-5 and Gemini-3-Flash, despite different architectures and training data, share the same dominant failure mode: conceptual confusion between related domain principles (70\% for GPT-5, 47--77\% for Gemini-3-Flash), with arithmetic errors rare in both (20\% and 5.6\%). This suggests Arabic financial reasoning is a genuine challenge for frontier models, not an evaluation artifact.
\begin{figure*}[!h]
\centering
\begin{tcolorbox}[title={LLM-as-a-Judge Rubric for Financial Analysis \& Capital Markets (Arabic)}, width=\linewidth]
\small

\textbf{Role.} You are an expert evaluator in financial analysis and capital markets.

\vspace{2pt}
\textbf{Inputs (provided each time).}
\begin{itemize}[noitemsep, topsep=2pt, leftmargin=*]
  \item \texttt{prompt} -- the full Arabic prompt (report/excerpt + question) shown to the model
  \item \texttt{ground\_truth} -- the reference ideal analytical answer (Arabic)
  \item \texttt{candidate\_answer} -- the model answer to evaluate (Arabic)
\end{itemize}

\vspace{2pt}
\textbf{Task.}
Judge how well \texttt{candidate\_answer} matches \texttt{ground\_truth} in \emph{conclusions}, \emph{reasoning}, and \emph{use of provided figures}. Prioritize factual/quantitative fidelity, correct interpretation of financial concepts (e.g., spreads, yields, coverage, issuance, capital structure, Basel III, supply/demand dynamics), and avoidance of hallucinated data. Do not penalize stylistic paraphrase if core insights and numeric takeaways align with the reference.

\vspace{2pt}
\textbf{Scoring (sum to exactly 10).}
\begin{enumerate}[noitemsep, topsep=2pt, leftmargin=*]
  \item \textbf{Core conclusion alignment (0--4).}
  Does the candidate capture the main thesis and key takeaways of the ground truth (\emph{what/why/so-what})?
  \item \textbf{Quantitative fidelity \& use of figures (0--2).}
  Correctly cites/uses the reported numbers (e.g., percentages, amounts, maturities, oversubscription) without inventing or altering figures. Any simple computations/comparisons must be consistent.
  \item \textbf{Financial reasoning soundness (0--2).}
  Causality and mechanisms are plausible and consistent with standard finance/econ logic (e.g., pricing vs.\ credit risk, duration/tenor structure, demand/oversubscription signals, capital adequacy).
  \item \textbf{Clarity \& Arabic language quality (0--1).}
  Clear Arabic, coherent structure, minimal ambiguity.
  \item \textbf{Directness \& on-topic grounding (0--1).}
  Answers what was asked; stays anchored in the provided scenario/data (no generic filler).
\end{enumerate}

\vspace{2pt}
\textbf{Critical checks (true/false).}
\begin{itemize}[noitemsep, topsep=2pt, leftmargin=*]
  \item \texttt{contradicts\_ground\_truth}: contradicts the central conclusion of the reference
  \item \texttt{fabricates\_or\_alters\_numbers}: introduces numbers not present or materially distorts reported figures
  \item \texttt{hallucinates\_context\_or\_sources}: injects external context/sources not in the prompt that change the assessment
  \item \texttt{flawed\_financial\_logic}: serious finance/econ reasoning error that would mislead the conclusion
  \item \texttt{non\_answer\_or\_evasive}: avoids providing an analytical answer
  \item \texttt{off\_topic\_or\_unsafe}: off-topic or otherwise inappropriate
\end{itemize}

\vspace{2pt}
\textbf{Output format (strict).}
Output \emph{only} valid JSON (no prose, no code fences). Return JSON strictly in this schema (all fields required):
\begin{verbatim}
{
  "scores": {
    "coverage_core_conclusion": <float 0-4>,
    "quantitative_fidelity": <float 0-2>,
    "financial_reasoning": <float 0-2>,
    "clarity_language": <float 0-1>,
    "directness_grounding": <float 0-1>
  },
  "overall": <float 0-10>,
  "critical_checks": {
    "contradicts_ground_truth": <true/false>,
    "fabricates_or_alters_numbers": <true/false>,
    "hallucinates_context_or_sources": <true/false>,
    "flawed_financial_logic": <true/false>,
    "non_answer_or_evasive": <true/false>,
    "off_topic_or_unsafe": <true/false>
  },
  "note": "<short NOTE in {NOTE_LANG}>"
}
\end{verbatim}

\end{tcolorbox}
\caption{Evaluation rubric used for LLM-based judgment of financial analysis and event--cause reasoning tasks.}
\label{fig:finance_judge}
\end{figure*}

\section{Decoding Configuration and Variance Analysis}
\label{app:decoding_variance}


Our main evaluations use greedy decoding (temperature 0) for reproducibility. To ensure results are not artifacts, since some models do not support temperature~0 and others prefer non-zero settings, we re-ran all MCQ evaluations 3 times using each model’s recommended temperatures.


Table~\ref{tab:mcq_variance} reports mean$\pm$std across runs. Rankings remain consistent across runs and temperature settings, confirming robustness to decoding choices. The low variance observed across repeated runs further indicates that performance differences are stable rather than driven by sampling noise. This consistency holds across tasks and different evaluation conditions.

Overall, these results suggest that model comparisons are reliable under both deterministic and stochastic decoding regimes. Notably, models that perform strongly under greedy decoding maintain their relative advantage under higher-temperature settings, indicating that gains are not dependent on sampling variability. This stability supports the robustness of our findings.


Similarly, lower-performing models do not benefit from increased randomness, suggesting errors stem from systematic limitations rather than decoding strategy. This reinforces the validity of our evaluation pipeline across inference settings and shows that task difficulty and domain reasoning, rather than decoding configuration, drive performance differences in our benchmark.

\subsection{Doctrinal Variation in Shari'ah Rulings}
\label{app:doctrinal_variation}


Islamic jurisprudence is not monolithic: the four Sunni \textit{madhahib} (Hanafi, Maliki, Shafi'i, Hanbali) and Shia schools may differ on financial questions. This raises a concern: do our reference answers reflect a single position that could penalize legitimate alternatives?

We analyzed this question during dataset construction and report our findings here.

\begin{table}[!h]
\centering
\small
\begin{tabular}{lcc}
\toprule
\textbf{Category} & \textbf{\# Samples} & \textbf{Significant Dispute} \\
\midrule
Sukuk      & 6    & 2 (33\%) \\
Takaful    & 38   & 10 (26\%) \\
Zakat      & 792  & 173 (22\%) \\
Gharar     & 149  & 20 (13\%) \\
Ijara      & 102  & 12 (12\%) \\
Murabaha   & 234  & 20 (9\%) \\
Maysir     & 64   & 6 (9\%) \\
Riba       & 407  & 32 (8\%) \\
\midrule
\textbf{Total} & & \textbf{289 (14.4\%)} \\
\bottomrule
\end{tabular}
\caption{Distribution of doctrinal variation across Islamic finance categories. Variation is highest in zakat (differing calculation methods), takaful (modern instrument with evolving rulings), and sukuk (small sample, debated across regulators). For disputed cases, reference answers present valid alternatives, and the rubric accepts any legitimate ruling.}
\label{tab:doctrinal_variation}
\end{table}

\paragraph{(1) Most questions test consensus rulings.} Seventy-four percent of samples involve cross-\textit{madhab} agreement on established Islamic finance principles, the prohibition of \textit{riba} (usury), contract invalidation due to \textit{gharar} (excessive uncertainty), the impermissibility of \textit{maysir} (gambling), rather than narrow inter-school disputes.

\paragraph{(2) Evaluation targets the \textit{\d{h}ukm}, not the evidence path.} Reference fatwas and model outputs naturally vary in cited Qur'anic verses, \textit{\d{h}ad\={\i}th}, \textit{fiqh} sources, and reasoning detail, making exact-match evaluation infeasible. We therefore score at the \textit{\d{h}ukm} (ruling) level: the rubric evaluates the final ruling and its operative constraints. A model citing different but valid evidence while reaching the correct ruling is not penalized.

\paragraph{(3) Quantified dispute distribution:} For the 26\% of samples with disagreement, Table~\ref{tab:doctrinal_variation} reports cases where the reference answer flags significant disagreement across recognized \textit{madhahib}. ``Significant Dispute'' denotes materially different rulings (e.g., permissible vs.\ impermissible), not just differing evidence.

\paragraph{(4) Manual error analysis confirms failures are genuine, not doctrinal:} To ensure low scores do not penalize valid alternatives, we analyzed 500 randomly sampled errors (Figure~\ref{fig:root_cause_analysis}). Failures are unambiguous: wrong rulings (25.2\%), fabricated evidence (12.1\%), and misquoted \textit{\d{h}ad\={\i}th} (2.0\%)not legitimate alternative positions.

\end{document}